\documentclass{article} 
\usepackage{iclr,times}
\iclrpreprint
\ClearShipoutPicture
\renewcommand{\headrulewidth}{0pt}

\usepackage[utf8]{inputenc} 
\usepackage[T1]{fontenc}
\usepackage{inconsolata}

\usepackage{hyperref}       
\usepackage{url}            
\usepackage{booktabs}       
\usepackage{amsfonts}       
\usepackage{nicefrac}       
\usepackage{microtype}      
\usepackage{xcolor}         

\usepackage[ruled,vlined]{algorithm2e}
\usepackage{amssymb}
\usepackage{amsmath}
\usepackage{amsthm}
\usepackage{xspace}
\usepackage{multirow}
\usepackage{graphicx}
\usepackage{comment}
\usepackage{subcaption}
\usepackage{enumitem}

\newcommand{\oursShort}{\textsc{MAPS}}

\makeatletter
\DeclareRobustCommand\onedot{\futurelet\@let@token\@onedot}
\def\@onedot{\ifx\@let@token.\else.\null\fi\xspace}
 
\def\ie{i.e\onedot} 
\def\iid{i.i.d\onedot} 
\def\etc{etc\onedot} 
\def\wrt{w.r.t\onedot}

\title{Model-agnostic predictability quantification of multivariable time series}

\author{
    \makebox[\dimexpr\textwidth-2\tabcolsep\relax][c]{%
        \normalfont
        \begin{tabular}{c}
            \\
            \\
            \textbf{Yuxuan Shu} \qquad \textbf{Vasileios Lampos} \\[0.5em]
            Centre for Artificial Intelligence\\
            Department of Computer Science \\
            University College London\\
            \{\texttt{yuxuan.shu.22},~\texttt{v.lampos}\}\texttt{@ucl.ac.uk}
        \end{tabular}%
  }
}

\begin{document}

\maketitle

\begin{abstract}
Multivariable time series forecasting plays an important role in various domains, where predictions depend on complex interactions among covariates. Forecasting models are commonly evaluated across numerous benchmark datasets. However, forecasting accuracy reflects both model capability and task predictability, making it difficult to characterise the difficulty of a task independently of the model used. Existing predictability measures often fail to capture nonlinear dependencies or are limited to univariate series. In this work, we propose \textbf{M}ultiv\textbf{a}riable \textbf{P}redictability \textbf{S}core (\oursShort{}), a model-agnostic metric for measuring the intrinsic predictability of multivariable time series forecasting tasks. \oursShort{} uses covariate information to assess predictability across different lookback windows and forecasting horizons, and provides a quantitative assessment of the predictive information available for forecasting. We evaluate~\oursShort{} across synthetic and real-world forecasting datasets against $6$ baseline predictability measures. The former confirms that our method closely aligns with the expected reference predictability ($r=0.963$). Empirical results further corroborate that \oursShort{} yields more reliable projections in the majority of experimental configurations. 

{
\smallskip
\textbf{Source code:} \href{https://github.com/ClaudiaShu/MAPS}{\nolinkurl{github.com/ClaudiaShu/MAPS}}
\vspace{0.2in}
}

\end{abstract}

\section{Introduction}
Time-series forecasting plays an important role across a wide range of domains and supports decision-making across multidisciplinary applications. Most forecasting problems are inherently multivariable. A single target variable is predicted from a set of covariates such as sensor information~\citep{gneiting2006calibrated,hinman2013solar}, economic indicators~\citep{asims1980macroeconomics,hstock2002forecasting,hstock2002macroeconomic}, or clinical measurements~\citep{lehman2014physiological,henderson2000joint,swulfsohn1997joint}. However, forecasting tasks differ considerably in the predictive information they contain.
Some exhibit strong temporal regularities and stable relationships between covariates and targets, whereas others are dominated by noise or weak predictive signals. This motivates a quantitative assessment of task predictability prior to modelling. Such an assessment can indicate whether a dataset supports reliable forecasting at a given horizon and provide insight into the performance that may be achievable.

Prior studies have attempted to characterise predictability in time series. An early approach~\citep{sugihara1994nonlinear} uses a forecasting method that predicts target values from similar historical patterns and relies on the forecasting accuracy to distinguish structured dynamics from stochastic noise. \citet{gholamzadeh2025towards} use mutual information to measure the relationship between inputs and targets and how much of this relationship remains in a model's prediction errors. \citet{beck2025mind} assess predictability by testing whether forecasting models consistently outperform a na\"ive persistence model, considering significant reductions in forecast error as evidence of predictable structure in the data. However, these approaches are model-dependent, meaning that the resulting predictability estimate depends on both the data and the forecasting model rather than reflecting the intrinsic aleatoric property of the prediction task. Early approaches from information theory deploy entropy to assess predictability.
Approximate entropy~\citep{mpincus1991approximate} measures the extent to which recurring patterns in a time series remain similar, where a lower value indicates that the series is more predictable. Permutation entropy~\citep{bandt2002permutation} uses Shannon entropy to measure the irregularity of changes in a time series, with lower values indicating higher predictability. However, neither method is formulated for multivariable forecasting tasks or considers how predictability changes across different forecasting horizons. Alternative approaches utilised the frequency domain. \citet{epowell1979spectral} introduced spectral entropy, which measures how consistently a time series follows repeating cycles, with lower values indicating clearer patterns and greater predictability. Building on this, ForeCA~\citep{goerg2013forecastable} extends to multivariate settings by forming weighted sums of variables and identifying those with lower spectral entropy as more predictable. More recently, \citet{wang2025spectral} transformed time series into the frequency domain and used distances between their frequency components as an indicator for forecasting model selection. Despite their computational efficiency, these frequency-domain approaches either consider a univariate setup~\citep{epowell1979spectral,wang2025spectral}, or do not account for nonlinear dependencies~\citep{goerg2013forecastable}.

Some prior work accounts for the forecast horizon when assessing predictability. \citet{wang2025accuracy} measures how much frequency patterns vary across time-series segments, including both the observed history and the target to be predicted, with greater variation indicating lower predictability. Similarly, \citet{catt2026limits} assesses predictability at each forecast horizon using the mutual information between the observed past and the target value, with lower mutual information indicating less predictive information at that horizon. However, these approaches are designed for univariate time series and do not account for covariate dependencies. Furthermore, mutual information estimation in high-dimensional spaces is known to be computationally prohibitive~\citep{kraskov2004estimating}, with complexity scaling poorly as the number of variables increases~\citep{gao2015efficient} (Figure~\ref{fig:runtime_comparison}).

Overall, existing predictability measures either depend on forecasting models, ignore forecasting horizons, or are limited to univariate settings. This gap is particularly relevant to the evaluation of forecasting models across benchmarks. Common long-term forecasting evaluations repeatedly use a small collection of datasets, including ETT, Electricity, Traffic, Weather, Exchange-Rate, and ILI~\citep{zeng2023are}, while broader archives have mostly focused on collections of univariate series~\citep{godahewa2021monash}. Recent benchmarks such as ProbTS~\citep{zhang2024probts}, GIFT-Eval~\citep{aksu2024gifteval}, and fev-bench~\citep{shchur2025fev-bench} have expanded evaluation across domains and multivariate settings. Nevertheless, benchmark tasks can still differ widely in the predictive information they contain. Some exhibit strong recurring patterns that can be captured from the target history, whereas others rely on cross-variable relationships or contain only weak predictive signals. A model-independent measure that considers the available covariates, lookback window, and forecast horizon can help interpret performance beyond conventional forecasting metrics.
    
Motivated by the aforementioned observations, our main contributions are:

\begin{enumerate}[leftmargin=*, topsep=-2.9pt, itemsep=-2.5pt]

\item We propose \textbf{M}ultiv\textbf{a}riable \textbf{P}redictability \textbf{S}core (\oursShort{}), a model-agnostic measure that quantifies the predictability of a multivariable forecasting task by assessing whether similar historical samples lead to consistent future target vectors. \oursShort{} explicitly accounts for the available covariates, lookback window, and forecasting horizon without requiring a forecasting model.

\item We evaluate \oursShort{} on synthetic time series with diverse dynamics and known predictability. \oursShort{} closely follows the estimated predictability, achieving an average linear correlation of $r=0.963$.

\item We compare \oursShort{} against $5$ existing predictability measures on multivariable Weather and Influenza-Like Illness forecasting tasks across multiple lookback windows and forecasting horizons. \oursShort{} shows the strongest relationship with empirical forecasting performance, obtaining correlations of $r=0.961$ on Weather and $r=0.827$ on ILI.\footnote{We note that these correlations are measuring different relationships which stem from the different experimental setups for these two forecasting tasks (see Section~\ref{subsec:pred_vs_forecasting}).} 

\end{enumerate}

{
\setlength{\belowcaptionskip}{-10pt}

\begin{figure}[!t]
    \centering
    \includegraphics[width=1.0\linewidth]{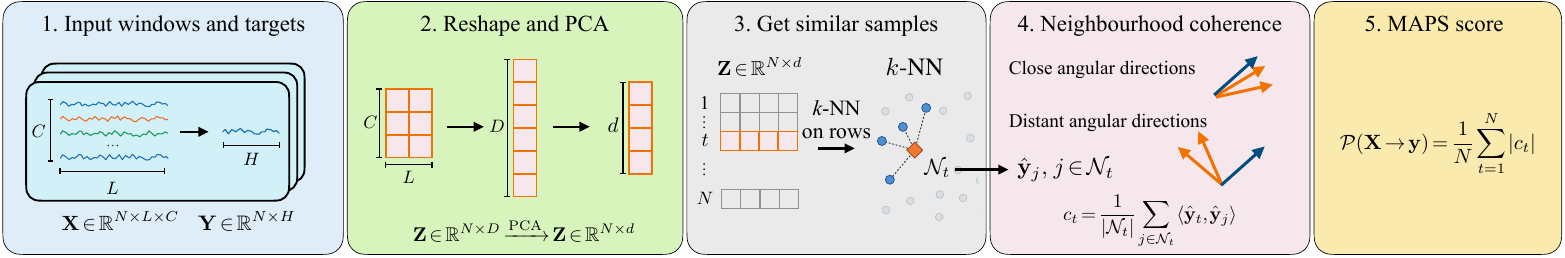}
    \caption{Overview of the proposed Multivariable Predictability Score (\oursShort{}).}
    \label{fig:method_overview}
\end{figure}
}

\section{Predictability of Multivariable Time Series -- Task Definition}
\label{sec:task}
We focus on quantifying the predictability of datasets for multivariable forecasting in a model-agnostic fashion. Specifically, we consider the multivariable forecasting task over a data set $\mathbf{Q}\in\mathbb{R}^{V \times C}$ where $V$ denotes the number of observations and $C$ is number of covariates, including $1$ target and $C-1$ exogenous variables. From $\mathbf{Q}$ we extract $N$ samples with a fixed stride of $M$. This forms an input tensor $\mathbf{X}\in\mathbb{R}^{N\times L\times C}$, where $L$ is the lookback window length. The corresponding future target values are captured in $\mathbf{Y}\in\mathbb{R}^{N\times H}$, where $H$ is the forecasting horizon. Each sample thus consists of an input $\mathbf{X}_t\!\in\!\mathbb{R}^{L\!\times\!C}$ covering the past time steps $\{t-L+1,\dots,t\}$, and the corresponding target vector $\mathbf{y}_t\!\in\!\mathbb{R}^H$ covering the target time steps $\{t+1,\dots,t+H\}$. The task is to define a predictability function $\mathcal{P}(\mathbf{X} \rightarrow \mathbf{Y})$ that quantifies the intrinsic predictability of the future target values from the corresponding input windows, independently of any specific model or supervised learning procedure.

\section{Multivariable Predictability Score (MAPS)}
\label{sec:predictability}
An overview of the~\oursShort{} method is shown in~\figureautorefname~\ref{fig:method_overview}. We first standardise each variable (or column) of $\mathbf{Q}$  over its full time series. We then partition the standardised time series of each variable into $N$ overlapping windows as described in section~\ref{sec:task}.
$\mathbf{X}_t$ and $\mathbf{y}_t$ are additionally standardised to maintain local patterns within each sample. 

\textbf{Dimensionality Reduction with Principal Component Analysis (PCA).} Following standardisation, each input window $\mathbf{X}_t$ is reshaped into vector $\mathbf{z}_t \in \mathbb{R}^{D}$, where $D = L \times C$. This forms the input state matrix held in $\mathbf{Z} \in \mathbb{R}^{N \times D}$. We note that the subsequent predictability computation operates on this $D$-dimensional representation and does not explicitly consider temporal adjacency. For multivariable datasets with many covariates or long lookback windows, $D$ can be rather large. To reduce the dimensionality of this representation, we optionally apply PCA~\citep{jolliffe1986principal} to project $\mathbf{Z}$ onto $d = \min(d, N-1, D)$ leading principal components, where $d$ is an adjustable target dimensionality. For simplicity, we use $\mathbf{Z} \in \mathbb{R}^{N \times d}$ and $\mathbf{z}_t \in \mathbb{R}^{d}$ to denote the matrix and state vectors after this optional projection, respectively, with $d = D$ when PCA is not applied.

\textbf{Identify Similar States with Nearest Neighbours.} We use $k$-nearest neighbours ($k$-NN) to identify similar samples among the $N$ sliced windows $\mathbf{Z}$. For each input sample $\mathbf{z}_t$, we identify its $k$ nearest neighbours $\mathcal{N}_t$ using the Euclidean distance, excluding $\mathbf{z}_t$ itself. The neighbourhood size $k$ determines a local component for the proposed predictability score. It should be large enough to provide statistically stable variance estimates, while small enough to preserve the local nature of the analysis. Following established practices in nearest-neighbour~\citep{jstone1977consistent,devroye1994strong,levada2024adaptive}, we set $k = \lfloor \log_2(N) \rfloor$.

\subsection{Quantifying Dataset Predictability}
\label{subsec:data_pred}

Our method is motivated by local regularity in the relationship between observed states and their future evolution~\citep{jstone1977consistent}. In deterministic dynamical systems, the current state determines the future trajectory~\citep{nlorenz1963deterministic}, while state-space reconstruction provides a basis for studying how nearby observations evolve~\citep{takens1981detecting,eckmann1985ergodic}. Accordingly, when the input contains predictive information about the target, nearby samples should correspond to target vectors with similar directions. We thus assess how consistently neighbouring samples align in the vector direction as an empirical measure of the predictive structure captured by the input samples.

Given the standardised target vectors ${\mathbf{y}_1,\ldots,\mathbf{y}_N}\in\mathbb{R}^{H}$,
we first obtain the corresponding unit direction vectors as $\hat{\mathbf{y}}_t = \frac{\mathbf{y}_t}{\lVert \mathbf{y}_t \rVert + \epsilon}$, where $\epsilon = 10^{-8}$ is a small constant that prevents division by zero. The resulting target vectors thus have unit length. The neighbourhood coherence between the current window $t$ and a neighbouring window $j$ is then obtained by computing the cosine similarity $\langle \hat{\mathbf{y}}_t, \hat{\mathbf{y}}_j \rangle \in [-1, 1]$. Averaging over $\mathcal{N}_t$, we have:
\begin{equation}
    c_t = \frac{1}{|\mathcal{N}_t|} \sum_{j \in \mathcal{N}_t} \left\langle \hat{\mathbf{y}}_t, \hat{\mathbf{y}}_j \right\rangle \, ,
\end{equation}
where $c_t\in[-1,1]$ measures the average directional agreement between window $t$ and the futures of windows in $\mathcal{N}_t$, whose observed histories are most similar.

The predictability of a time series at window $t$ is then given by $s_t = \lvert c_t \rvert$, where, by construction, $s_t\in[0,1]$. Importantly, we take the absolute value of $c_t$, after averaging the directional similarities. This is because both positive ($c_t \rightarrow 1$) and negative ($c_t \rightarrow -1$) directional agreement of the target vectors are equally indicative of predictability. In this case, large $\lvert c_t \rvert$ values indicate that the $k$-nearest neighbours of window $t$ follow coherent future directions, reflecting a more predictable mapping between similar historical samples and their corresponding target vectors. By contrast, neighbourhoods containing orthogonal future directions yield $c_t \approx 0$, indicating that windows with similar observed histories do not evolve consistently. In such cases, the input representation provides limited information about future dynamics.

The overall predictability score of a time series dataset \wrt the specific look-back window and forecasting horizon is thus given by:
\begin{equation}
    \mathcal{P}(\mathbf{X} \rightarrow \mathbf{y})
    = \frac{1}{N} \sum_{t=1}^{N} s_t \in [0, 1]
    \, \, \, .
\end{equation}
Intuitively, $\mathcal{P}(\mathbf{X} \rightarrow \mathbf{y}) \approx 1$ indicates high predictability, where samples that are close in the input space have futures that also evolve similarly, indicating a more deterministic input-to-target mapping. Contrarily, $\mathcal{P}(\mathbf{X} \rightarrow \mathbf{y}) \approx 0$ indicates low predictability, \ie even nearly identical samples result in diverged directions, suggesting chaotic dynamics that are potentially less predictable.

\textbf{Computation cost.}
Performing PCA cost $O(ND\min\{N,D\})$, an exhaustive Euclidean nearest-neighbour search over $N$ windows costs $O(N^2d)$, and conducting cosine similarity over the neighbouring target vectors costs $O(NkH)$. Given $d$ and $H$ are fixed and $k=\lfloor\log_2 N\rfloor$, this complexity simplifies to $O(ND\min\{N,D\}+N^2)$.
We additionally provide runtime comparison in~\appendixautorefname~\ref{appsubsec:runtime}.

\noindent\textbf{Constant target variable.} A special case emerges when the target variable is a constant. Standardisation is applied over the entire sequence, so a constant target will become equal to zero in every window ($\mathbf{y}_t = 0$) and leave the direction undefined. Given that a constant target can be forecast exactly by its own value, we set $s_t = 1$ regardless. A near-constant target is otherwise scored normally as it does not collapse to zero vector. We further discuss this in~\appendixautorefname~\ref{appsubsec:synthetic_results}.

\section{Results}
We evaluate \oursShort{} on both synthetic and real-world datasets and compare it with $6$ baseline methods. We further examine its behaviour under distribution shift and robustness to different configurations.

\subsection{Multivariable Time Series Datasets}

\noindent\textbf{Synthetic time series.} To evaluate our method under controlled conditions, we construct synthetic datasets with controllable predictability. We begin by combining signals with known deterministic and stochastic components, including sinusoidal, linear trends, exponential and logarithmic growth, and Gaussian noise. By varying the proportion and magnitude of each component, we control the level of predictability. In this way, we form $6$ basic synthetic data sets. We further consider $6$ complex dynamic systems to enable a more comprehensive evaluation: a linear Gaussian state-space model, a smooth Gaussian process, a fractional Gaussian noise process, a two-state hidden Markov model, a noisy Lorenz system, and a quasi-periodic process. These cover a range of different predictability decay behaviours \wrt the forecasting horizon, enabling a horizon-dependent evaluation of our predictability metric~\citep{dong2025time}.

\noindent\textbf{Real-world tasks.} We form the benchmark multivariable time series datasets from the Weather (\texttt{WEA}) and Influenza-Like Illness (\texttt{ILI}) forecasting tasks introduced by~\cite{shu2026sonnet}. For \texttt{WEA}, the forecasting target is variable `t850' at a target location (for $20$ locations in total), and the time step is set to $6$ hours. We consider both a task that uses $5$ covariates at the target location, and also expand this to a task with $45$ covariates across a $3\times3$ grid of locations (target and 8 neighbouring ones). \texttt{ILI} contains daily ILI rates and corresponding Google search frequencies for related queries for USA HHS Regions 2, 9, and 10, and England, UK. 

Further details for both synthetic and real-world datasets are provided in~\appendixautorefname~\ref{appsec:datasets_metrics}.

\subsection{Experiment configurations}
\label{subsec:experiment_configurations}

\textbf{Baseline methods.}  We compare \oursShort{} against $5$ model-agnostic baseline predictability methods. These are a persistence baseline, ForeCA~\citep{goerg2013forecastable}, Spectral Entropy~\citep{wang2025spectral}, Pattern Complexity (PC)~\citep{wang2025accuracy}, and Mutual Information (MI). To adapt baselines that are originally designed for univariate forecasting to multivariable settings, we compute the scores for each input variable \wrt the target and average across variables. We additionally include Epiplexity~\citep{finzi2026entropy}, a model-dependent measurement, where we use ridge regression as the predictive model in our experiments. Further details are provided in Appendix~\ref{appsec:supplementary_exp}.

\textbf{Experiment setup.} We conduct experiments with varying lookback window lengths ($L$) and forecasting horizons ($H$). Specifically, we consider $L\!\in\!\{28,56,84\}$, $H\!\in\!\{4,12,28\}$ for \texttt{WEA}, and $L\!\in\!\{28,56\}$, $H\!\in\!\{7,14,21,28\}$ for \texttt{ILI}. We obtain $d=50$ principal components and set the neighbourhood size to $k=\lfloor\log_2 N\rfloor$. Unless otherwise stated, we sample rolling windows using a stride of $M=10$.

\textbf{Evaluation metrics.} We use the linear correlation coefficient $r$ between estimated and reference predictability scores as the primary metric in our evaluation. For synthetic data, the reference value is the designed predictability of each data-generating process. For real-world data, we use the performance of a competitive machine learning model as the reference. Machine learning performance is mainly determined by the linear correlation coefficient $\rho$ between the model predictions and observed targets, a metric that is invariant to both scaling and translation. We additionally consider a scaled absolute forecasting error as a complementary measure. Specifically, let $e$ denote the MAE of the model on the test set, and $\sigma_y$ be the standard deviation of the target series over the test period. We first compute the standardised MAE $e/\sigma_y$~\citep{gasteiger2020directional}, which expresses absolute forecasting error in units of target standard deviation. We further transform this into the scaled MAE, $s=\sigma_y/(\sigma_y+e)\in(0, 1]$, with higher values indicating better forecasting performance. 

\textbf{Forecasting configuration.} For the controlled synthetic systems, we use ridge regression as an interpretable forecasting baseline, with the penalty coefficient set to $1$. For the real-world experiments, we use the performance of multivariable forecasting models as an empirical reference to evaluate the predictability scores. Predictability is computed on the training set, and forecasting performance is evaluated on the test set. Training settings and the Weather and ILI data splits are detailed in Appendix~\ref{appsec:supplementary_exp}.

\subsection{Predictability under Controlled Conditions with Synthetic Data Sets}

\begin{figure}
    \centering
    \includegraphics[width=0.9\linewidth]{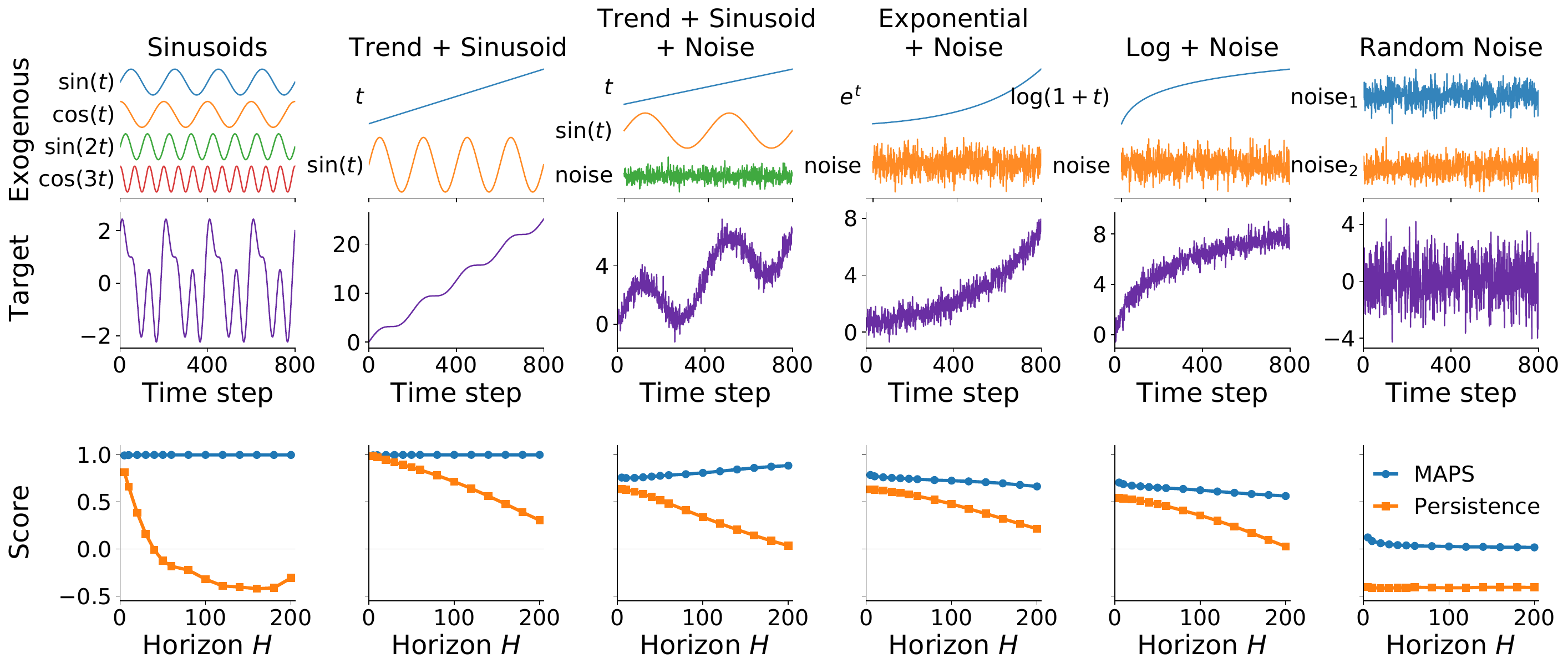}
    \caption{Basic synthetic time series used to validate~\oursShort{}. The exogenous variables are shown in the top row, the corresponding target series in the middle row, and the predictability \wrt $H$ in the bottom row, each is a combination of the exogenous inputs. Predictability scores are computed using a rolling window where the input includes all covariates, \ie both exogenous variables and the target.}
    \label{fig:simple_synthetic_data}
\end{figure}

We commence by conducting experiments using $6$ basic synthetic data sets over $800$ time steps with known predictability and consider a forecasting setting of $L=50$. The first two are deterministic, and their expected predictability score is 1. The $3-5$ series adds \iid Gaussian noise to $3$ deterministic functions, making the series less predictable. The sixth is only Gaussian noise and is therefore not predictable. Outcomes are depicted in \figureautorefname~\ref{fig:simple_synthetic_data}. \oursShort{} assigns a score of $1$ to both deterministic series across different $H$, about $0.77$ to the series with added noise, and close to $0$ to pure Gaussian noise, corroborating the expected predictability. Contrarily, persistence reflects the pattern of the series rather than its predictability.
\begin{figure}[t]
    \centering
    \includegraphics[width=0.99\linewidth]{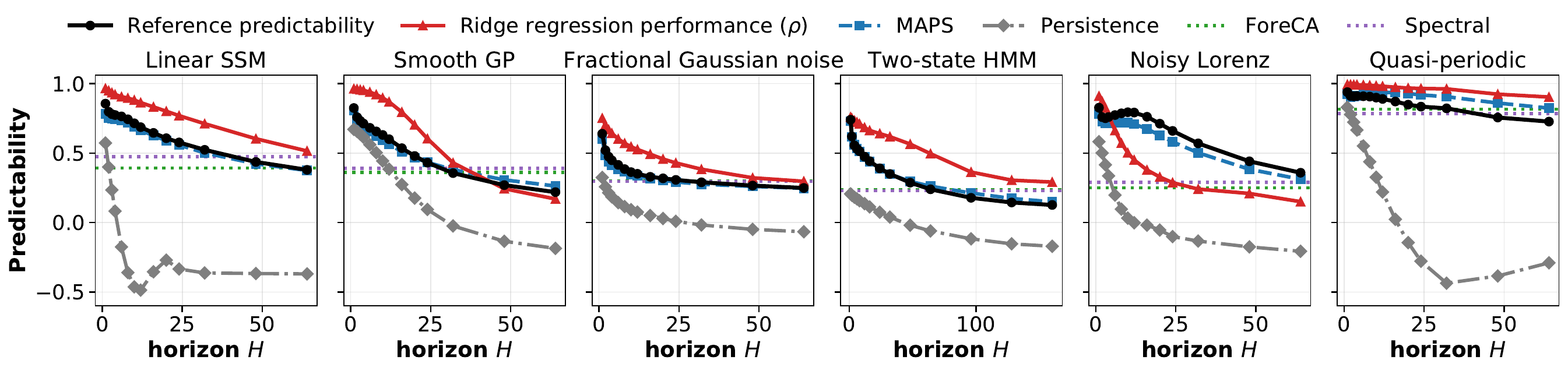}
    \caption{Reference predictability compared to~\oursShort{}, ForeCA, Spectral, Persistence, and the empirical performance (using linear correlation) of a ridge regression model for complex synthetic time series tasks across increasing forecasting horizons ($H$) for a fixed lookback window ($L=16$).}
    \label{fig:chaos_horizon}
\end{figure}

We further evaluate \oursShort{} on $6$ synthetic univariate tasks that exhibit more complex and nonlinear characteristics and hence have different degrees of predictability. This univariate setting ($C=1$) serves as a special case of the forecasting task considered by \oursShort{}. For each task, we generate sequences of $4{,}000$ time steps and compare a reference (expected) predictability score with \oursShort{} \wrt different forecasting horizons ($H$), under a fixed look-back window of $L=16$. We define the reference predictability as
\begin{equation}
P(L,H) = 1- \frac{\mathbb{E}\!\left[\sum_{h=1}^{H}\operatorname{Var}(y_{t+h}\mid y_{t-L+1:t})\right]}{\sum_{h=1}^{H}\operatorname{Var}(y_{t+h})} \, \, ,
\end{equation}
where $y_t$ is the generated future target value at time $t$. We use the known data-generating process to obtain possible future sequences given the observations in the lookback window and calculate how much uncertainty remains across these futures. $P(L,H) \in [0,1]$ quantifies uncertainty after observing past values within the look-back window $L$. A value of $1$ indicates that the past fully determines the target value, while $0$ dictates that the past provides no predictive information (see also Appendix~\ref{appsubsubsec:true_pred}). 

Figure~\ref{fig:chaos_horizon} compares the reference predictability with~\oursShort{}, $2$ frequency-based baselines (ForeCA and Spectral), persistence baseline, alongside the average correlation between predictions and ground truth for a ridge regression model as a baseline. It can be observed that \oursShort{} strongly correlates with $P(L,H)$ ($r=.963$). Their scores actually overlap, with the exception of the quasi-periodic task where MAPS slightly overestimates predictability. Ridge regression performance estimates exhibit slightly weaker agreement with the target predictability than MAPS, but still a significant one ($r=.924$). This alignment further motivates the use of forecasting model performance as the reference predictability in the follow-up experiments. Persistence achieves a lower correlation ($r=.852$) and oftentimes underestimates predictability, particularly at longer forecasting horizons. Given this mediocre performance and the fact that its score reflects the error of the persistence predictor rather than data predictability, we do not report Persistence in the following experiments. Additionally, the two spectral baselines perform poorly on datasets without strong seasonality and fail to capture how predictability varies across different $L/H$ settings.

\begin{figure}[!t]
    \centering
    \includegraphics[width=0.95\linewidth]{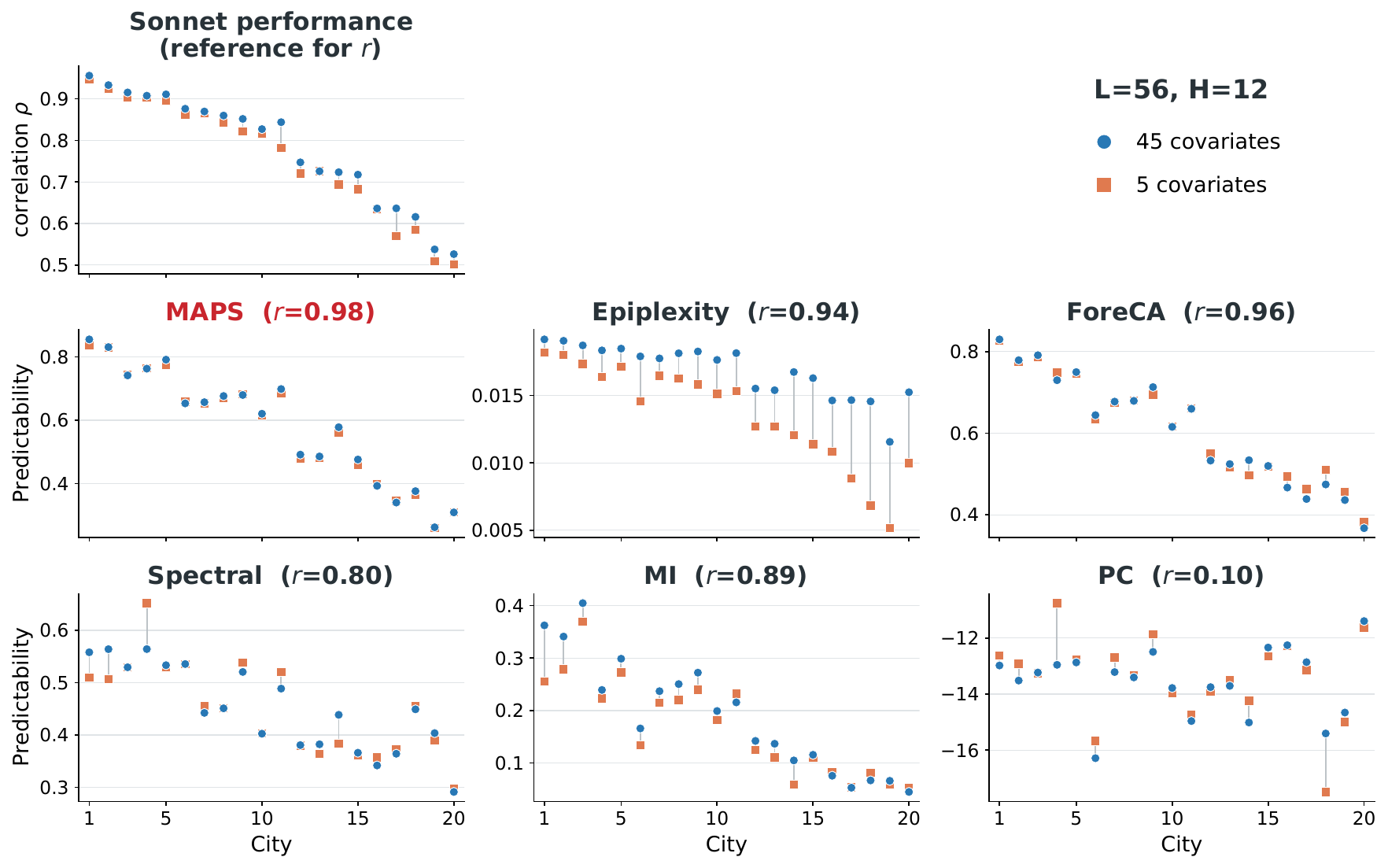}
    \caption{Predictability scores of~\oursShort{} and $5$ baselines for the \texttt{WEA} task compared against the forecasting performance of Sonnet across $20$ locations (lookback $L{=}56$, horizon $H{=}12$) for two configurations ($5$ or $45$ covariates). Location order reflects Sonnet's correlation performance.}
    \label{fig:pred_score_corr}
\end{figure}

\subsection{Evaluating Predictability Against Empirical Forecasting Performance}
\label{subsec:pred_vs_forecasting}

We further evaluate predictability measurements on real-world data sets (\texttt{WEA} and \texttt{ILI}). For \texttt{WEA}, we first average predictability and performance scores across the test periods and then compute their correlation over $20$ locations for each forecasting horizon. We perform this for two configurations, considering just the target location ($5$ covariates) or including a grid of $8$ neighbouring locations ($45$ covariates). Results are shown in~\figureautorefname~\ref{fig:pred_score_corr} and Table~\ref{tab:weather_sonnet_correlations}. Across the compared methods, \oursShort{} shows the strongest correlation with the performance ($\rho$) of Sonnet. Among the baselines, Epiplexity and ForeCA are the most competitive. However, Epiplexity is model-dependent, meaning that its effectiveness is closely related to the model used.

Additionally, including climate indicators from neighbouring regions (45 covariates) is expected to improve the reference performance $\rho$ over using only local indicators (5 covariates) by providing additional predictive information. Importantly, \oursShort{} reflects this improvement. The configuration with more covariates yields higher \oursShort{} scores. The baseline measures largely fail to capture this distinction, except for the model-based method Epiplexity and MI.

\begin{table*}[!b]
\centering
\small
\setlength{\tabcolsep}{3.0pt}
\caption{Average linear correlation (across all test periods) between predictability scores and Sonnet's forecasting performance, which is measured by correlation ($\rho$) or scaled MAE ($s$). Correlation breakdowns for lookback windows ($L$), forecasting horizons ($H$), and covariate number ($C$) are provided for the \texttt{WEA} task, over the $20$ considered locations. 
The right-most two columns enumerate the average correlation among all combinations of $(L,H)$ across regions for \texttt{WEA} and \texttt{ILI}.
}
\label{tab:weather_sonnet_correlations}
\resizebox{\textwidth}{!}{%
\begin{tabular}{cl*{20}{c}}
\toprule
& \multirow{3}{*}{Method}
& \multicolumn{6}{c}{\texttt{WEA}, $L=28$}
& \multicolumn{6}{c}{\texttt{WEA}, $L=56$}
& \multicolumn{6}{c}{\texttt{WEA}, $L=84$}
& \multicolumn{2}{c}{Across $(L,H)$ pairs} \\
\cmidrule(lr){3-8}
\cmidrule(lr){9-14}
\cmidrule(lr){15-20}
\cmidrule(lr){21-22}
&
& \multicolumn{2}{c}{$H\!=\!4$}
& \multicolumn{2}{c}{$H\!=\!12$}
& \multicolumn{2}{c}{$H\!=\!28$}
& \multicolumn{2}{c}{$H\!=\!4$}
& \multicolumn{2}{c}{$H\!=\!12$}
& \multicolumn{2}{c}{$H\!=\!28$}
& \multicolumn{2}{c}{$H\!=\!4$}
& \multicolumn{2}{c}{$H\!=\!12$}
& \multicolumn{2}{c}{$H\!=\!28$}
& \multirow{2}{*}{\texttt{WEA}}
& \multirow{2}{*}{\texttt{ILI}} \\
\cmidrule(lr){3-4}
\cmidrule(lr){5-6}
\cmidrule(lr){7-8}
\cmidrule(lr){9-10}
\cmidrule(lr){11-12}
\cmidrule(lr){13-14}
\cmidrule(lr){15-16}
\cmidrule(lr){17-18}
\cmidrule(lr){19-20}
&
& $C\!=\!45$ & $C\!=\!5$
& $C\!=\!45$ & $C\!=\!5$
& $C\!=\!45$ & $C\!=\!5$
& $C\!=\!45$ & $C\!=\!5$
& $C\!=\!45$ & $C\!=\!5$
& $C\!=\!45$ & $C\!=\!5$
& $C\!=\!45$ & $C\!=\!5$
& $C\!=\!45$ & $C\!=\!5$
& $C\!=\!45$ & $C\!=\!5$
& & \\
\midrule

\multirow{6}{*}{\rotatebox[origin=c]{90}{$\rho$}}
& \oursShort{}
& .843 & .896 & \textbf{.975} & .971
& .967 & .972
& .812 & .872 & \textbf{.977} & \textbf{.975}
& \textbf{.972} & \textbf{.968}
& .797 & .856 & \textbf{.975} & \textbf{.973}
& \textbf{.973} & \textbf{.976}
& \textbf{.961} & \textbf{.827} \\

& Epiplexity
& \textbf{.991} & \textbf{.955} & .951 & .951
& \textbf{.968} & \textbf{.976}
& \textbf{.987} & \textbf{.956} & .922 & .953
& .940 & .967
& \textbf{.961} & \textbf{.959} & .853 & .949
& .859 & .974
& .833 & .455 \\

& ForeCA
& .801 & .836 & .972 & \textbf{.974}
& .961 & .972
& .804 & .845 & .976 & .974
& .970 & .963
& .805 & .850 & .974 & .972
& \textbf{.973} & .973
& -- & -- \\

& Spectral
& .459 & .514 & .844 & .768
& .851 & .779
& .471 & .525 & .836 & .763
& .840 & .780
& .467 & .526 & .836 & .779
& .848 & .788
& -- & -- \\

& MI
& .741 & .780 & .902 & .868
& .906 & .883
& .718 & .766 & .910 & .873
& .914 & .881
& .711 & .757 & .905 & .878
& .928 & .897
& .470 & .403 \\

& PC
& .350 & .458 & .036 & .328 & -.031 & .309
& .297 & .436 & -.057 & .260 & -.089 & .237
& .292 & .452 & -.120 & .217 & -.101 & .237
& .371 & .335 \\

\midrule

\multirow{6}{*}{\rotatebox[origin=c]{90}{$s$}}
& \oursShort{}
& .921 & \textbf{.946} & \textbf{.960} & .951
& \textbf{.955} & \textbf{.952}
& .913 & \textbf{.939}
& \textbf{.969} & \textbf{.959}
& \textbf{.958} & \textbf{.961}
& .905 & .934
& \textbf{.964} & \textbf{.958}
& \textbf{.957} & \textbf{.960}
& \textbf{.946} & \textbf{.817} \\

& Epiplexity
& \textbf{.923} & .902 & .907 & .896
& .944 & .938
& .909 & .896 & .887 & .895
& .917 & .945
& .873 & .904 & .822 & .892
& .846 & .940
& .833 & .537 \\

& ForeCA
& .915 & .936 & .956 & \textbf{.960}
& .947 & \textbf{.952}
& \textbf{.922} & .938 & .965 & \textbf{.959}
& .953 & .955
& \textbf{.921} & \textbf{.940} & .958 & .956
& .953 & .954
& -- & -- \\

& Spectral
& .664 & .666 & .905 & .813
& .902 & .834
& .687 & .668 & .903 & .819
& .888 & .838
& .678 & .672 & .897 & .822
& .893 & .832
& -- & -- \\

& MI
& .899 & .888 & .907 & .861
& .901 & .871
& .895 & .885 & .931 & .878
& .920 & .891
& .886 & .874 & .927 & .876
& .931 & .901
& .450 & .481 \\

& PC
& .335 & .427 & .054 & .305 & -.038 & .271
& .215 & .332 & -.062 & .208 & -.097 & .208
& .168 & .322 & -.153 & .161 & -.130 & .177
& .357 & .267 \\

\bottomrule
\end{tabular}%
}
\vspace{0.3em}
\end{table*}

Results across all forecasting configurations of \texttt{WEA} are enumerated in Table~\ref{tab:weather_sonnet_correlations}. When we compare the predictability score with target correlation $\rho$, \oursShort{} yields the strongest correlation $r$ in 9 out of 18 configurations. It performs best in five of six configurations with $H=12$ and in four out of six configurations with $H=28$. Epiplexity is the most competitive baseline when referred to $\rho$, obtaining stronger results for the shortest forecasting horizon $(H=4)$. Turning to the scaled-MAE score $s$, \oursShort{} yields the strongest correlation in 13 out of 18 configurations. ForeCA is the most competitive baseline for $s$, achieving the highest correlation in six configurations, while Epiplexity leads in one. Notably, \oursShort{} is more competitive at longer forecasting horizons. It leads or ties in five of six settings at $H=12$ and four of six at $H=28$ for target correlation $\rho$, and in five and six settings, respectively, for scaled MAE. \oursShort{}'s alignment with $s$ is also significantly stronger than Epiplexity's (paired $t$-test, $p<0.001$). The final two columns provide a complementary within-location analysis across different $L, H$ settings. Baselines that do not produce task-specific scores (ForeCA and Spectral) are marked ``--'' because their scores remain constant. \oursShort{} achieves the highest within-location correlation in \texttt{WEA}/\texttt{ILI}, reaching $0.961/0.827$ with $\rho$, and $0.946/0.817$ with $s$. These results show that the alignment between \oursShort{} and forecasting performance is consistent across datasets and $(L, H)$ settings, capturing changes in forecasting difficulty across tasks.

\subsection{Predictability Under Distribution Shift}

\begin{figure*}[!t]
    \centering

    \begin{minipage}[t]{0.66\textwidth}
        \vspace{0pt}
        \centering
        \includegraphics[width=\linewidth]
            {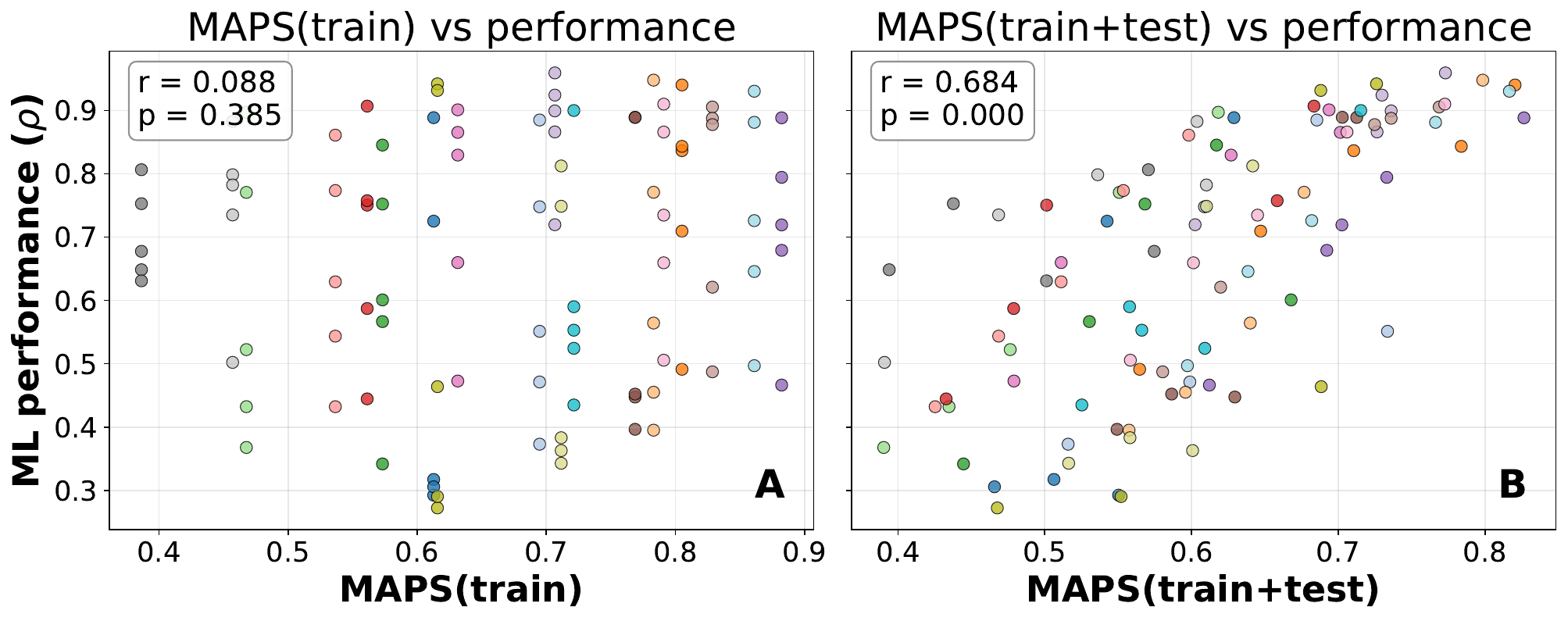}
        \caption{Correlation between predictability score and downstream forecasting performance across weather regions under distribution shift. (A) Predictability computed on the source (training) data. (B) Predictability computed on the combined training and test data.}
        \label{fig:distribution_shift_weather}
    \end{minipage}
    \hfill
    \begin{minipage}[t]{0.3\textwidth}
        \vspace{0pt}
        \centering
        \small
        \setlength{\tabcolsep}{3.2pt}
        \resizebox{\textwidth}{!}{        \begin{tabular}{lcc}
            \toprule
            Method & Train & Train+test \\
            \midrule
            MAPS        &  0.088 & 0.684 \\
            Epiplexity  &  0.110 & 0.534 \\
            MI          &  0.067 & 0.523 \\
            Spectral    & $-0.099$ & 0.531 \\
            ForeCA      &  0.108 & 0.519 \\
            PC          & $-0.109$ & 0.381 \\
            \bottomrule
        \end{tabular}
        }        \captionof{table}{Linear correlation $r$ between each predictability measure and the downstream forecasting performance across the same cross-location weather tasks.}
        \label{tab:distribution_shift_baselines}
    \end{minipage}
\end{figure*}

\begin{table*}[!b]
\centering
\captionsetup{
  justification=raggedright,
  singlelinecheck=false
}

\begin{minipage}[t]{0.36\textwidth}
\vspace{0pt}
\caption{Component ablation of MAPS. Correlations are averaged over all
combinations of $(L,H)$ across locations for \texttt{WEA} and \texttt{ILI}. Best results are \textbf{bolded}.}
\label{tab:ablation_study}

\centering
\resizebox{\linewidth}{!}{%
\begin{tabular}{@{}lcc@{}}
\toprule
Method & \texttt{WEA} ($r$) & \texttt{ILI} ($r$) \\
\midrule
MAPS
    & \textbf{0.961} & \textbf{0.827} \\
w/ variance
    & 0.882          & 0.593 \\
w/ signed directions
    & 0.902          & 0.480 \\
w/o standardisation
    & \textbf{0.961} & 0.733 \\
w/o window standardisation
    & \textbf{0.961} & 0.811 \\
w/o PCA
    & 0.960          & 0.607 \\
\bottomrule
\end{tabular}%
}
\end{minipage}
\hfill
\begin{minipage}[t]{0.62\textwidth}
\vspace{0pt}
\caption{MAPS predictability scores for England ILI under different
PCA variance thresholds for each $(L,H)$ task. ``No PCA'' uses all
input variables, while the other rows retain the fewest principal components
needed to explain $70\%$, $80\%$, or $90\%$ of the variance.}
% . The highest score in each column is \textbf{bolded}.}
\label{tab:pca_variance_ili_eng}

\centering
\resizebox{\linewidth}{!}{%
\begin{tabular}{@{}lrrrrrrrr@{}}
\toprule
& \multicolumn{4}{c}{$L=28$}
& \multicolumn{4}{c}{$L=56$} \\
\cmidrule(lr){2-5}
\cmidrule(lr){6-9}
PCA
& $H\!=\!7$
& $H\!=\!14$
& $H\!=\!21$
& $H\!=\!28$
& $H\!=\!7$
& $H\!=\!14$
& $H\!=\!21$
& $H\!=\!28$ \\
\midrule
No PCA
& 0.470 & 0.467 & 0.463 & 0.458
& 0.541 & 0.517 & 0.510 & 0.497 \\
70\%
& 0.578 & 0.568 & 0.561 & 0.556
& 0.607 & 0.598 & 0.593 & 0.589 \\
80\%
& 0.558 & 0.542 & 0.531 & 0.525
& 0.581 & 0.573 & 0.570 & 0.568 \\
90\%
& 0.552 & 0.564 & 0.567 & 0.572
& 0.619 & 0.638
& 0.648 & 0.653 \\
\bottomrule
\end{tabular}%
}
\end{minipage}
\end{table*}

We evaluate whether dataset predictability provides a useful prior for forecasting performance under real-world distribution shift. Specifically, we train a forecasting model on one region’s weather data and evaluate it on another, constructing a diverse set of cross-location transfer tasks to approximate the distribution shift.
To ensure a comprehensive evaluation, we train the forecasting model on each of the 20 locations and evaluate it on five other locations, selecting the two geographically nearest and three geographically farthest locations for each case.

Figure~\ref{fig:distribution_shift_weather}.A shows that the predictability score of the training data alone shows a weak relationship with downstream performance ($r=0.088$). This is expected as the source dataset provides limited information about the predictability of the target series if there is a distribution shift. In contrast, when predictability is estimated on the combined training and deployment data (Figure~\ref{fig:distribution_shift_weather}.B), a slightly more positive correlation emerges ($r=0.684$). Table~\ref{tab:distribution_shift_baselines} also shows that other baselines present a similar change. It can be concluded that the current predictability scores measure the intrinsic predictability of the time series on which they are computed, and they do so accurately once that distribution is observed. However, it should be noted that in deployment settings subject to shift, a score computed at training time would overstate or understate the difficulty of the task.

\subsection{Ablation study for \oursShort{}}
\label{subsec:ablation}

In this section, we perform an ablation study by replacing one component of~\oursShort{} at a time to justify our design choice. `w/ variance' replaces the direction-based similarity measure with a variance-ratio objective using the mean variance of the target and neighbouring features. Further detail in~\ref{appsubsec:abltaion_components}. `w/ signed directions' is instead of taking the absolute value, we map $[-1,1]$ directly onto $[0,1]$, so a negative alignment is mapped below $0.5$. `w/o standardisation' removes the z-score standardisation applied before window slicing. `w/o window standardisation' removes the window-wise standardisation for each sliced window. Finally, `w/o PCA' omits the PCA projection before the nearest-neighbour search. Table~\ref{tab:ablation_study} shows how each design choice performs under controlled predictability scenarios and contributes to \oursShort{}. \oursShort{} score consistently achieves the strongest alignment with downstream forecasting performance on the real-world datasets, obtaining the highest correlation on both \texttt{WEA} ($r=0.961$) and \texttt{ILI} ($r=0.827$). Removing the standardisation has little effect, while removing PCA substantially degrades performance alignment on the \texttt{ILI} dataset (the dataset that contains more covariates), suggesting that dimensionality reduction improves the reliability of neighbourhood estimation in higher-dimensional input spaces. Replacing the neighbourhood direction alignment with a variance measurement or using signed directions also reduces agreement with forecasting performance.

\subsection{Sensitivity analysis}

\begin{figure}[!t]
    \centering
    \includegraphics[width=0.89\linewidth]{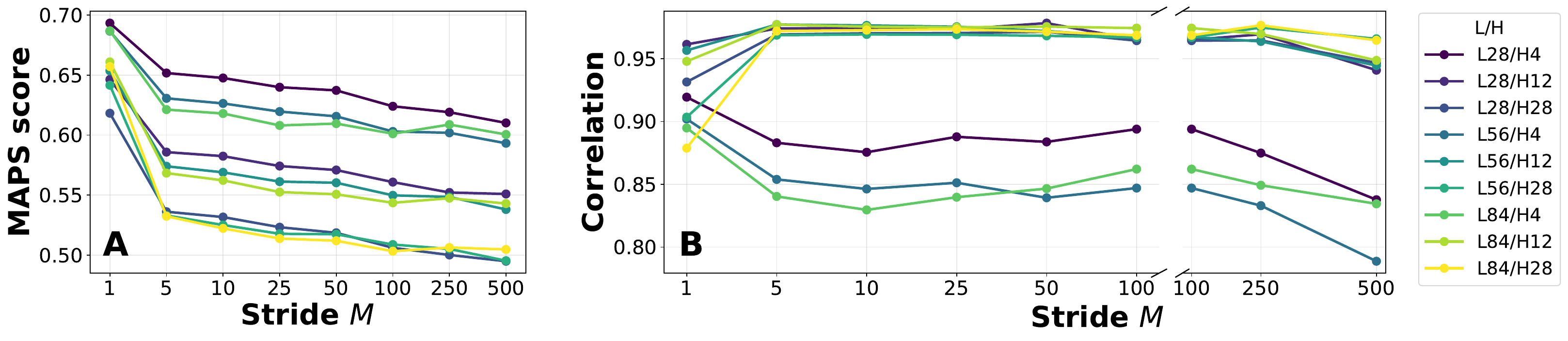}
    \vspace{-5pt}
    \caption{Effect of window stride on \oursShort{} across nine $(L,H)$ settings across $20$ locations (\texttt{WEA}). (A) Averaged predictability score; (B) Correlation across $20$ locations.
    The axis breaks at stride 100, which separates the equally spaced stride values from a linear scale over 100--500.}
    \label{fig:stride_effectiveness}
\end{figure}

\textbf{Effect of PCA dimensionality.} In \oursShort{}, PCA is applied to remove redundancy among all available variables, thereby reducing the computational cost of the following $k$-nearest-neighbour search. This is particularly important for the \texttt{ILI} dataset, where each dataset contains $2\text{k}$–$5\text{k}$ covariates. Table~\ref{tab:pca_variance_ili_eng} reports the \oursShort{} score after retaining different proportions of the variance. The score is consistently higher after PCA than the no PCA version. We note that these results present as a sensitivity analysis of \oursShort{} to the input. They do not, by themselves, show that a higher score is more accurate.

\textbf{Effect of window stride.} As shown in Figure~\ref{fig:stride_effectiveness}.A, increasing the stride ($M$) lowers the absolute predictability score because adjacent windows overlap less, making neighbouring samples less similar. Despite this change in score magnitude, its correlation with downstream forecasting performance remains stable across a wide range of strides (Figure~\ref{fig:stride_effectiveness}.B). Stride 1 performs slightly worse, as heavily overlapping windows may be selected as neighbours rather than revealing recurring patterns. Correlation begins to drop at larger strides (250 and 500), where too few windows remain to accurately capture the local structure of the dataset. Arguably, other factors such as dataset size and seasonality may also contribute to this decline, though we leave a closer analysis to future work. Overall, \oursShort{} is robust to the choice of stride.

\section{Conclusion}

We propose \oursShort{}, a model-agnostic measure for quantifying the predictability of a multivariable time-series forecasting task given its covariates, lookback window, and forecasting horizon. Across synthetic systems with known predictability, \oursShort{} closely follows the reference predictability, achieving an average linear correlation of $r=0.963$. Furthermore, across Weather and Influenza-Like Illness forecasting tasks, \oursShort{} shows the strongest relationship with empirical forecasting performance among the evaluated predictability measures, obtaining correlations of $r=0.961$ and $r=0.827$, respectively. Overall, \oursShort{} provides a complementary measure for characterising forecasting tasks and interpreting model performance across heterogeneous datasets.

\subsection*{Acknowledgments}
V.~Lampos would like to thank all levels of support from grant EP/X031276/1 (EPSRC) and the SOFAIR Lab (UKRI).

\bibliography{main}

@article{gholamzadeh2025towards,
    title        = {{Towards Measuring Predictability: To which extent data-driven approaches can extract deterministic relations from data exemplified with time series prediction and classification}},
    author       = {Saleh Gholam Zadeh and Vaisakh Shaj and Patrick Jahnke and Gerhard Neumann and Tim Breitenbach},
    year         = 2025,
    journal      = {Transactions on Machine Learning Research},
}

@article{wang2025accuracy,
    title        = {{Accuracy law for the future of deep time series forecasting}},
    author       = {Wang, Yuxuan and Wu, Haixu and Ma, Yuezhou and Fang, Yuchen and Zhang, Ziyi and Liu, Yong and Wang, Shiyu and Ye, Zhou and Xiang, Yang and Wang, Jianmin and others},
    year         = 2025,
    journal      = {arXiv preprint arXiv:2510.02729}
}

@article{beck2025mind,
    title        = {{Mind the naive forecast! a rigorous evaluation of forecasting models for time series with low predictability}},
    author       = {Beck, Nico and Dovern, Jonas and Vogl, Stefanie},
    year         = 2025,
    journal      = {Applied Intelligence},
    volume       = 55,
    number       = 6,
    pages        = 395
}

@article{mpincus1991approximate,
    title        = {{Approximate entropy as a measure of system complexity.}},
    author       = {S M Pincus},
    year         = 1991,
    journal      = {Proceedings of the National Academy of Sciences},
    volume       = 88,
    number       = 6,
    pages        = {2297--2301}
}

@article{shu2025deformtime,
    title        = {{DeformTime: capturing variable dependencies with deformable attention for time series forecasting}},
    author       = {Yuxuan Shu and Vasileios Lampos},
    year         = 2025,
    journal      = {Transactions on Machine Learning Research},
}

@article{nlorenz1963deterministic,
    title        = {{Deterministic Nonperiodic Flow}},
    author       = {Edward N.  Lorenz},
    year         = 1963,
    journal      = {Journal of Atmospheric Sciences},
    volume       = 20,
    number       = 2,
    pages        = {130--141}
}

@article{eckmann1985ergodic,
    title        = {{Ergodic theory of chaos and strange attractors}},
    author       = {Eckmann, J. -P. and Ruelle, D.},
    year         = 1985,
    journal      = {Rev. Mod. Phys.},
    volume       = 57,
    pages        = {617--656},
    issue        = 3,
    numpages     = {0}
}

@inproceedings{takens1981detecting,
    title        = {{Detecting strange attractors in turbulence}},
    author       = {Takens, Floris},
    year         = 1981,
    booktitle    = {Dynamical Systems and Turbulence, Warwick 1980},
    pages        = {366--381},
    isbn         = {978-3-540-38945-3}
}

@article{levada2024adaptive,
    title        = {{Adaptive $k$-nearest neighbor classifier based on the local estimation of the shape operator}},
    author       = {Levada, Alexandre Lu{\'\i}s Magalh{\~a}es and Nielsen, Frank and Haddad, Michel Ferreira Cardia},
    year         = 2024,
    journal      = {arXiv preprint arXiv:2409.05084}
}

@article{jstone1977consistent,
    title        = {{Consistent Nonparametric Regression}},
    author       = {Charles J. Stone},
    year         = 1977,
    journal      = {The Annals of Statistics},
    volume       = 5,
    number       = 4,
    pages        = {595--620},
}

@article{devroye1994strong,
    title        = {{On the Strong Universal Consistency of Nearest Neighbor Regression Function Estimates}},
    author       = {Luc Devroye and Laszlo Gyorfi and Adam Krzyzak and Gabor Lugosi},
    year         = 1994,
    journal      = {The Annals of Statistics},
    volume       = 22,
    number       = 3,
    pages        = {1371 -- 1385}
}

@article{finzi2026entropy,
    title        = {{From Entropy to Epiplexity: Rethinking Information for Computationally Bounded Intelligence}},
    author       = {Finzi, Marc and Qiu, Shikai and Jiang, Yiding and Izmailov, Pavel and Kolter, J Zico and Wilson, Andrew Gordon},
    year         = 2026,
    journal      = {arXiv preprint arXiv:2601.03220}
}

@article{wang2025spectral,
    title        = {{Spectral Predictability as a Fast Reliability Indicator for Time Series Forecasting Model Selection}},
    author       = {Wang, Oliver and Quan, Pengrui and Yang, Kang and Srivastava, Mani},
    year         = 2025,
    journal      = {arXiv preprint arXiv:2511.08884}
}

@article{shu2026sonnet,
    title        = {{Sonnet: Spectral Operator Neural Network for Multivariable Time Series Forecasting}},
    author       = {Yuxuan Shu and Vasileios Lampos},
    year         = 2026,
    journal      = {Proceedings of the AAAI Conference on Artificial Intelligence}
}

@article{shchur2025fev-bench,
    title        = {{fev-bench: A realistic benchmark for time series forecasting}},
    author       = {Shchur, Oleksandr and Ansari, Abdul Fatir and Turkmen, Caner and Stella, Lorenzo and Erickson, Nick and Guerron, Pablo and Bohlke-Schneider, Michael and Wang, Yuyang},
    year         = 2025,
    journal      = {arXiv preprint arXiv:2509.26468}
}

@inbook{jolliffe1986principal,
    title        = {{Principal Components in Regression Analysis}},
    author       = {Jolliffe, I. T.},
    year         = 1986,
    booktitle    = {Principal Component Analysis},
    pages        = {129--155},
}

@article{bandt2002permutation,
    title        = {{Permutation Entropy: A Natural Complexity Measure for Time Series}},
    author       = {Bandt, Christoph and Pompe, Bernd},
    year         = 2002,
    journal      = {Phys. Rev. Lett.},
    volume       = 88,
    pages        = 174102,
    issue        = 17,
    numpages     = 4
}

@article{epowell1979spectral,
    title        = {{A spectral entropy method for distinguishing regular and irregular motion of Hamiltonian systems}},
    author       = {G E Powell and I C Percival},
    year         = 1979,
    journal      = {Journal of Physics A: Mathematical and General},
    volume       = 12,
    number       = 11,
    pages        = 2053
}

@article{catt2026limits,
    title        = {{On the Limits of Prediction: Forecastability Profiles and Information Decay in Time Series}},
    author       = {Catt, Peter Maurice},
    year         = 2026,
    journal      = {arXiv preprint arXiv:2603.20546}
}

@article{sugihara1994nonlinear,
    title        = {{Nonlinear forecasting for the classification of natural time series}},
    author       = {Sugihara, George and Grenfell, Bryan Thomas and May, Robert McCredie and Tong, H.},
    year         = 1994,
    journal      = {Philosophical Transactions of the Royal Society of London, Series A: Physical and Engineering Sciences},
    volume       = 348,
    number       = 1688,
    pages        = {477--495}
}

@inproceedings{goerg2013forecastable,
    title        = {{Forecastable Component Analysis}},
    author       = {Goerg, Georg},
    year         = 2013,
    booktitle    = {Proceedings of the 30th International Conference on Machine Learning},
    series       = {Proceedings of Machine Learning Research},
    volume       = 28,
    number       = 2,
    pages        = {64--72}
}

@article{kraskov2004estimating,
    title        = {{Estimating mutual information}},
    author       = {Kraskov, Alexander and St\"ogbauer, Harald and Grassberger, Peter},
    year         = 2004,
    journal      = {Phys. Rev. E},
    volume       = 69,
    pages        = {066138},
}

@inproceedings{gao2015efficient,
    title        = {{Efficient Estimation of Mutual Information for Strongly Dependent Variables}},
    author       = {Gao, Shuyang and Ver Steeg, Greg and Galstyan, Aram},
    year         = 2015,
    booktitle    = {Proceedings of the Eighteenth International Conference on Artificial Intelligence and Statistics},
    series       = {Proceedings of Machine Learning Research},
    volume       = 38,
    pages        = {277--286},
}

@article{lehman2014physiological,
    title        = {{A physiological time series dynamics-based approach to patient monitoring and outcome prediction}},
    author       = {Lehman, Li-Wei H and Adams, Ryan P and Mayaud, Louis and Moody, George B and Malhotra, Atul and Mark, Roger G and Nemati, Shamim},
    year         = 2014,
    journal      = {IEEE J Biomed Health Inform},
    volume       = 19,
    number       = 3,
    pages        = {1068--1076},
}

@article{asims1980macroeconomics,
    title        = {{Macroeconomics and Reality}},
    author       = {Christopher A. Sims},
    year         = 1980,
    journal      = {Econometrica},
    publisher    = {[Wiley, Econometric Society]},
    volume       = 48,
    number       = 1,
    pages        = {1--48},
    issn         = {00129682, 14680262},
}

@article{hstock2002forecasting,
    title        = {{Forecasting Using Principal Components from a Large Number of Predictors}},
    author       = {James H. Stock and Mark W. Watson},
    year         = 2002,
    journal      = {Journal of the American Statistical Association},
    publisher    = {[American Statistical Association, Taylor & Francis, Ltd.]},
    volume       = 97,
    number       = 460,
    pages        = {1167--1179},
    issn         = {01621459},
}

@article{hstock2002macroeconomic,
    title        = {{Macroeconomic Forecasting Using Diffusion Indexes}},
    author       = {James H Stock and Mark W Watson},
    year         = 2002,
    journal      = {Journal of Business \& Economic Statistics},
    publisher    = {Taylor \& Francis},
    volume       = 20,
    number       = 2,
    pages        = {147--162},
}

@article{henderson2000joint,
    title        = {{Joint modelling of longitudinal measurements and event time data}},
    author       = {Henderson, R and Diggle, P and Dobson, A},
    year         = 2000,
    journal      = {Biostatistics},
    volume       = 1,
    number       = 4,
    pages        = {465--480},
}

@article{swulfsohn1997joint,
    title        = {{A Joint Model for Survival and Longitudinal Data Measured with Error}},
    author       = {Michael S. Wulfsohn and Anastasios A. Tsiatis},
    year         = 1997,
    journal      = {Biometrics},
    publisher    = {International Biometric Society},
    volume       = 53,
    number       = 1,
    pages        = {330--339},
    issn         = {0006341X, 15410420},
}

@article{gneiting2006calibrated,
    title        = {{Calibrated Probabilistic Forecasting at the Stateline Wind Energy Center}},
    author       = {Tilmann Gneiting and Kristin Larson and Kenneth Westrick and Marc G Genton and Eric Aldrich},
    year         = 2006,
    journal      = {Journal of the American Statistical Association},
    publisher    = {Taylor \& Francis},
    volume       = 101,
    number       = 475,
    pages        = {968--979},
}

@article{hinman2013solar,
    title        = {{Solar forecasting methods for renewable energy integration}},
    author       = {Rich H. Inman and Hugo T.C. Pedro and Carlos F.M. Coimbra},
    year         = 2013,
    journal      = {Progress in Energy and Combustion Science},
    volume       = 39,
    number       = 6,
    pages        = {535--576},
    issn         = {0360-1285}
}

@article{rasp2020weatherbench,
    title        = {{WeatherBench: a benchmark data set for data-driven weather forecasting}},
    author       = {Rasp, Stephan and Dueben, Peter D and Scher, Sebastian and Weyn, Jonathan A and Mouatadid, Soukayna and Thuerey, Nils},
    year         = {2020},
    journal      = {Journal of Advances in Modeling Earth Systems},
    volume       = {12},
    number       = {11},
    pages        = {e2020MS002203}
}

@inproceedings{zhang2023crossformer,
    title        = {{Crossformer: Transformer Utilizing Cross-Dimension Dependency for Multivariate Time Series Forecasting}},
    author       = {Zhang, Yunhao and Yan, Junchi},
    year         = {2023},
    booktitle    = {International Conference on Learning Representations},
    furl         = {https://openreview.net/forum?id=vSVLM2j9eie}
}

@inproceedings{zeng2023are,
    title        = {{Are Transformers Effective for Time Series Forecasting?}},
    author       = {Zeng, Ailing and Chen, Muxi and Zhang, Lei and Xu, Qiang},
    year         = {2023},
    booktitle    = {Proceedings of the AAAI Conference on Artificial Intelligence},
    pages        = {11121--11128}
}

@article{chang2004predictability,
    title        = {{Predictability of Linear Coupled Systems. Part I: Theoretical Analyses}},
    author       = {Ping Chang and R. Saravanan and Timothy DelSole and Faming Wang},
    year         = 2004,
    journal      = {Journal of Climate},
    volume       = 17,
    number       = 7,
    pages        = {1474--1486}
}

@article{park2011domain,
    title        = {{Domain Decomposition Approach for Fast Gaussian Process Regression of Large Spatial Data Sets}},
    author       = {Chiwoo Park and Jianhua Z. Huang and Yu Ding},
    year         = 2011,
    journal      = {Journal of Machine Learning Research},
    volume       = 12,
    number       = 47,
    pages        = {1697--1728}
}

@inproceedings{williams1995gaussian,
    title        = {{Gaussian Processes for Regression}},
    author       = {Williams, Christopher and Rasmussen, Carl},
    year         = 1995,
    booktitle    = {Advances in Neural Information Processing Systems},
    volume       = 8,
    pages        = {},
}

@article{terenin2024numerically,
    title        = {{Numerically Stable Sparse Gaussian Processes via Minimum Separation using Cover Trees}},
    author       = {Alexander Terenin and David R. Burt and Artem Artemev and Seth Flaxman and Mark van der Wilk and Carl Edward Rasmussen and Hong Ge},
    year         = 2024,
    journal      = {Journal of Machine Learning Research},
    volume       = 25,
    number       = 26,
    pages        = {1--36}
}

@article{ranjan2011computationally,
    title        = {{A Computationally Stable Approach to Gaussian Process Interpolation of Deterministic Computer Simulation Data}},
    author       = {Pritam Ranjan and Ronald Haynes and Richard Karsten},
    year         = 2011,
    journal      = {Technometrics},
    volume       = 53,
    number       = 4,
    pages        = {366--378}
}

@article{rabiner1989tutorial,
    title        = {{A tutorial on hidden Markov models and selected applications in speech recognition}},
    author       = {Rabiner, L.R.},
    year         = 1989,
    journal      = {Proceedings of the IEEE},
    volume       = 77,
    number       = 2,
    pages        = {257--286}
}

@article{wen2019gaussian,
    title        = {{A Gaussian mixture method for specific differential phase retrieval at {X}-band frequency}},
    author       = {Wen, G. and Fox, N. I. and Market, P. S.},
    year         = 2019,
    journal      = {Atmospheric Measurement Techniques},
    volume       = 12,
    pages        = {5613--5637}
}

@article{gordon1993novel,
    title        = {{Novel approach to nonlinear/non-Gaussian Bayesian state estimation}},
    author       = {N.J. Gordon  and D.J. Salmond  and A.F.M. Smith},
    year         = 1993,
    journal      = {IEE Proceedings F (Radar and Signal Processing)},
    volume       = 140,
    pages        = {107--113},
    issue        = 2
}

@article{kitagawa1996monte,
    title        = {{Monte Carlo Filter and Smoother for Non-Gaussian Nonlinear State Space Models}},
    author       = {Genshiro Kitagawa},
    year         = 1996,
    journal      = {Journal of Computational and Graphical Statistics},
    volume       = 5,
    number       = 1,
    pages        = {1--25}
}

@article{delsole2004predictability,
    title        = {{Predictability and Information Theory. Part I: Measures of Predictability}},
    author       = {Timothy DelSole},
    year         = 2004,
    journal      = {Journal of the Atmospheric Sciences},
    volume       = 61,
    number       = 20,
    pages        = {2425--2440}
}

@article{godahewa2021monash,
    title        = {{Monash time series forecasting archive}},
    author       = {Godahewa, Rakshitha and Bergmeir, Christoph and Webb, Geoffrey I and Hyndman, Rob J and Montero-Manso, Pablo},
    year         = 2021,
    journal      = {arXiv preprint arXiv:2105.06643}
}

@inproceedings{zhang2024probts,
    title        = {{ProbTS: Benchmarking Point and Distributional Forecasting across Diverse Prediction Horizons}},
    author       = {Zhang, Jiawen and Wen, Xumeng and Zhang, Zhenwei and Zheng, Shun and Li, Jia and Bian, Jiang},
    year         = 2024,
    booktitle    = {Advances in Neural Information Processing Systems},
    volume       = 37,
    pages        = {48045--48082}
}

@article{aksu2024gifteval,
    title        = {{Gift-eval: A benchmark for general time series forecasting model evaluation}},
    author       = {Aksu, Taha and Woo, Gerald and Liu, Juncheng and Liu, Xu and Liu, Chenghao and Savarese, Silvio and Xiong, Caiming and Sahoo, Doyen},
    year         = 2024,
    journal      = {arXiv preprint arXiv:2410.10393}
}

@inproceedings{gasteiger2020directional,
    title        = {{Directional Message Passing for Molecular Graphs}},
    author       = {Johannes Gasteiger and Janek Groß and Stephan Günnemann},
    year         = 2020,
    booktitle    = {International Conference on Learning Representations}
}

@article{dong2025time,
    title        = {{Time-lagged recurrence: A data-driven method to estimate the predictability of dynamical systems}},
    author       = {Dong, Chenyu and Faranda, Davide and Gualandi, Adriano and Lucarini, Valerio and Mengaldo, Gianmarco},
    year         = 2025,
    journal      = {Proceedings of the National Academy of Sciences},
    volume       = 122,
    number       = 20,
    pages        = {e2420252122}
}
\bibliographystyle{abbrvnat}

\newpage

\appendix
\setcounter{table}{0}
\renewcommand{\thetable}{S\arabic{table}}\setcounter{figure}{0}
\renewcommand{\thefigure}{S\arabic{figure}}\setcounter{equation}{0}
\renewcommand{\theequation}{S\arabic{equation}}

\section*{Appendix}
\addcontentsline{toc}{section}{Appendix}

\section{Datasets and Metrics}
\label{appsec:datasets_metrics}

\subsection{Choice of model performance metrics}
\label{appsubsec:metrics_choice}
\begin{figure}[b]
    \centering
    \includegraphics[width=0.89\linewidth]{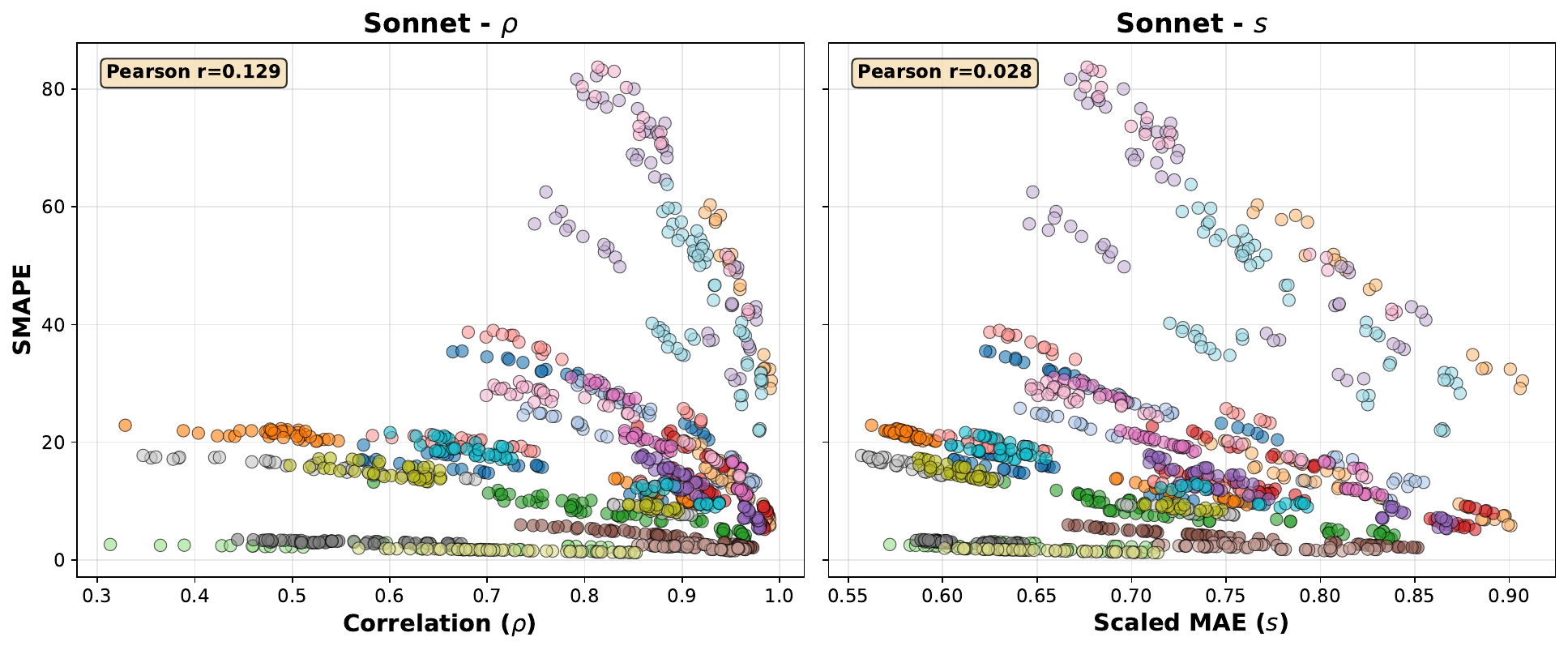}
    \caption{Relationship between SMAPE and linear correlation across weather forecasting tasks for Sonnet against target correlation $\rho$ (left) and scaled MAE score $s$ (right). Each point represents one task for one of 20 locations, with colours indicating different locations.}
    \label{appfig:ml_metric_ranking}
\end{figure}

The proposed \oursShort{} score is designed to quantify the intrinsic predictability of a time series under a rolling-window setting in a model-agnostic way. As described in Section~\ref{subsec:experiment_configurations}, for real-world datasets, where reference predictability is difficult to quantify directly, we assess the score by comparing it with the performance of a forecasting model on the same task under the metrics of linear correlation ($\rho$) and scaled MAE ($s$). Notably, the SMAPE score of ML models has recently been used to benchmark predictability scores~\citep{wang2025spectral}. In Figure~\ref{appfig:ml_metric_ranking}, we compare SMAPE with linear correlation ($\rho$) and scaled MAE ($s$) for Sonnet~\citep{shu2026sonnet}, where each location is shown in a different colour. Within each location, both $\rho$ and $s$ show a negative relationship with SMAPE as expected. Models with higher correlation tend to have lower prediction errors under the same data distribution. However, this relationship becomes weaker when results are evaluated across locations. This suggests that these metrics capture different aspects of forecasting performance: a predictability score that correlates well with $\rho$ and $s$ need not correlate well with SMAPE. For completeness, we additionally report the relationship between the proposed predictability score and target SMAPE in Appendix~\ref{appsubsec:results_mae_smape}. 

\subsection{Generation of Complex Synthetic Datasets with Known Predictability}
\label{appsubsec:gen_synthetic_data}

\begin{figure}[t]
    \centering
    \includegraphics[width=0.99\linewidth]{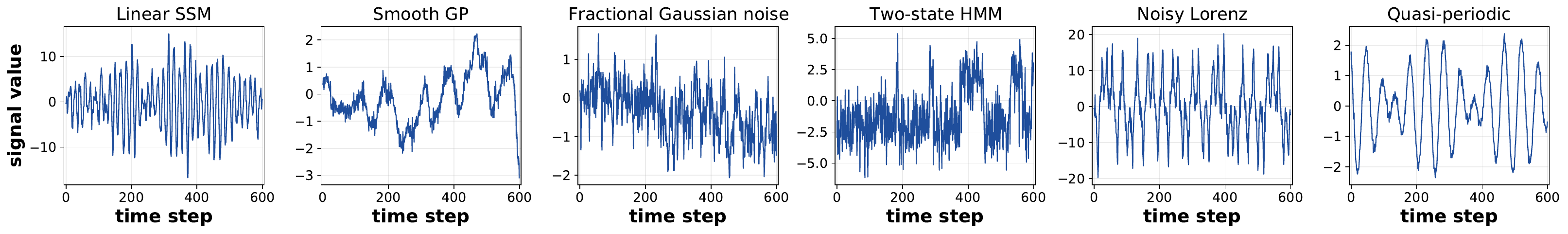}
    \caption{Generated time series for the six synthetic processes with known predictability. Each series contains $T=4{,}000$ time steps, of which the first $600$ are shown.}
    \label{appfig:plots_synthetic}
\end{figure}

We construct six synthetic time series with known predictability to evaluate the proposed method under controlled conditions. \figureautorefname~\ref{appfig:plots_synthetic} shows the generated series. For the linear Gaussian state-space model (SSM), smooth Gaussian Process (GP), fractional Gaussian noise, and quasi-periodic process, we characterise temporal dependence through autocovariance functions. Throughout this section, $y_t\in\mathbb{R}$ denotes the observed univariate series at time $t$, and $\gamma(h) = \operatorname{Cov}(y_t,y_{t+h})\in\mathbb{R}$ denotes its autocovariance function at lag $h$, which does not depend on $t$ because the process is stationary.

\textbf{Linear Gaussian state-space model.} We first generate synthetic time series using a $2$-dimensional linear Gaussian state-space model. Let $z_t=(z_{t,1},z_{t,2})^\top\in\mathbb R^2$ be the latent state at time $t$. The state transition is given by
\begin{equation}
    z_{t+1} = A z_t + w_t,
    \qquad
    y_t = z_{t,1} + v_t \, \, ,
\end{equation}
where $w_t \sim \mathcal{N}(0,\sigma_w^2I_2)$ and $v_t \sim \mathcal{N}(0,\sigma_v^2)$ are independent Gaussian noise. The transition matrix is defined as
\begin{equation}
    A = \kappa
    \begin{pmatrix}
        \cos(\omega) & -\sin(\omega) \\
        \sin(\omega) & \phantom{-}\cos(\omega)
    \end{pmatrix},
    \qquad
    \omega = 2\pi f \, \, ,
\end{equation}
where $f$ controls the frequency and $\kappa < 1$ controls the rate at which the process loses temporal dependence. This construction produces a stationary, stochastically perturbed oscillatory process with exponentially decaying temporal dependence, while $\sigma_v^2$ controls the noise level. In our experiments, we set $f=0.06$, $\kappa=0.985$, $\sigma_w^2=1$, and $\sigma_v^2=0.35$, and initialise with $z_0=(0,0)^\top$.

\textbf{Smooth Gaussian Process.}
The smooth GP is defined by the squared-exponential covariance function, given by:
\begin{equation}
    \gamma(h)=
    \begin{cases}
    1+\sigma_\epsilon^2 & h=0\\
    \displaystyle\exp\left(-\frac{h^2}{2\ell^2}\right) & h\neq0 \, \, ,
    \end{cases}
\end{equation}
where $h$ denotes the temporal lag between two time points. $\ell$ controls the rate at which temporal correlations decay with increasing time lag, while $\sigma_\epsilon^2$ controls the variance of the noise, which comes into play only when \(h=0\). In our experiments, we set $\ell=12$ and $\sigma_\epsilon^2=0.05$.

\textbf{Fractional Gaussian noise process.}
The next synthetic time series is generated using fractional Gaussian noise, a zero-mean stationary GP with covariance function
\begin{equation}
    \gamma(h)=
    \begin{cases}
        1+\sigma_\epsilon^2 & h=0\\
        \displaystyle
        \frac{1}{2}
        \left[
            (h+1)^{2\eta}
            -2h^{2\eta}
            +|h-1|^{2\eta}
        \right] & h\neq0 \, \, ,
    \end{cases}
\end{equation}
where $h$ denotes the temporal lag and $\eta\in(0,1)$ controls how much a time series depends on the past values. For $\eta>1/2$, the process exhibits long-range dependence, with temporal correlations reducing according to a power law. Larger values of $\eta$ therefore produce stronger and more persistent dependence across long time lags. The term $\sigma_\epsilon^2$ is the noise variance. We set $\eta=0.95$ and $\sigma_\epsilon^2=0.01$.

\textbf{Two-state hidden Markov model.} The next series is generated by a two-state hidden Markov model with Gaussian emissions. At each time point $t$, the process is assumed to be in one of two possible states, denoted by $z_t\in\{1,2\}$. The transition probabilities are given by
\begin{equation}
    \Pr(z_t=j\mid z_{t-1}=i) = A_{ij},
    \qquad
    A =
    \begin{pmatrix}
        \lambda & 1-\lambda\\
        1-\lambda & \lambda
    \end{pmatrix} \, \, ,
\end{equation}
where $\Pr(\cdot)$ denotes probability, $A_{ij}$ is the probability of transitioning from state $i$ at time $t-1$ to state $j$ at time $t$, and $\lambda$ is the probability of remaining in the same state. Given the current state, the time series is generated from a state-specific Gaussian distribution:
\begin{equation}
    y_t \mid z_t=i \sim \mathcal{N}(\mu_i,\sigma_i^2) \, \, ,
\end{equation}
where $\mu_i$ is the mean of state $i$, and $\sigma_i$ is its standard deviation. The two hidden states have their own mean and standard deviation, allowing the series to switch between two regimes. We set $\lambda=0.99$, $(\mu_1,\mu_2)=(-2,2)$, and $(\sigma_1,\sigma_2)=(1.4,1.4)$.

\textbf{Noisy Lorenz system.} This series is generated through a stochastic Lorenz state-space model. Let $z_t\in\mathbb{R}^3$ denote the latent state, which evolves according to
\begin{equation}
    z_{t+1} = \Phi_{\mathrm{RK4}}(z_t) + w_t,
    \qquad
    y_t = z_{t,1} + v_t \, \, ,
\end{equation}
where $z_{t+1}$ is the state at the next time point, $\Phi_{\mathrm{RK4}}$ is the function that takes the current state $z_t$ and computes its next state according to the Lorenz differential equations using the fourth-order Runge–Kutta (RK4) method, $w_t\sim\mathcal{N}(0,\sigma_w^2 I_3)$ is process noise that introduces randomness into the state evolution, $z_{t,1}$ is the first component of the latent state, and $v_t\sim\mathcal{N}(0,\sigma_v^2)$ is the added noise. We use the standard Lorenz parameters $(\sigma,\rho,\beta)=(10,28,8/3)$, an integration step of $\Delta t=0.05$, and noise variance $\sigma_w^2=0.02$ and $\sigma_v^2=4$. We initialise with $z_0=(1,1,1)^\top$.

\textbf{Quasi-periodic process.}
We next combine several sinusoidal components with frequencies that do not have a common period. This yields
\begin{equation}
    y_t =
    \sum_{j=1}^{J} a_j
    \cos(2\pi f_j t + \phi_j)
    + \epsilon_t \, \, ,
\end{equation}
where $J$ is the number of sinusoidal components, $a_j$ is the amplitude of component $j$, $f_j$ is its frequency, $\phi_j$ is its fixed phase, and $\epsilon_t\sim\mathcal{N}(0,\sigma_\epsilon^2)$ is independent noise. Because the frequencies do not share a common period, the combined signal does not exactly repeat over time. The underlying signal is fully deterministic because the amplitudes, frequencies, and phases are fixed, while the added noise can make the observed series appear irregular. We set $J=2$, $(a_1,a_2)=(1,0.7)$, $f_1=0.017$, $f_2=(\sqrt{5}-1)f_1$, and $\sigma_\epsilon^2=0.01$.

\subsubsection{Reference Predictability}
\label{appsubsubsec:true_pred}

Following \citet{chang2004predictability}, we define reference predictability as one minus the variance that remains across possible future sequences given the lookback window, normalised by the total variance in the future target values. For a lookback window of $L$ time steps and a forecasting horizon $H$, the reference predictability is given by
\begin{equation}
P(L,H)
=
1-
\frac{
\mathbb{E}\!\left[
\sum_{h=1}^{H}\operatorname{Var}
(y_{t+h}\mid y_{t-L+1:t})
\right]
}{
\sum_{h=1}^{H}\operatorname{Var}(y_{t+h})
} \, \, .
\end{equation}
For each past input window, we compute how much the possible future target values can vary under the known data-generating process. The variance is computed directly for the $4$ GP-based and the hidden Markov models, and approximated with generated futures for the noisy Lorenz system. The reference predictability score is then computed by $1$ minus the normalised variance averaged across the sliced windows. $P(L,H)=1$ thus indicates a completely predictable target, whereas $P(L,H)=0$ means that observing the past provides no information about the target value. For each synthetic system, we first compute its conditional predictive variance. For the linear Gaussian SSM, smooth GP, fractional Gaussian noise, and quasi-periodic Gaussian process, the conditional variance is available analytically from the autocovariance function $\gamma(h)$~\citep{williams1995gaussian,park2011domain}. Let
\begin{equation}
    \Sigma_{pq}=\gamma(|p-q|),
    \qquad
    c_p=\gamma(H+L-1-p) \, \, ,
\end{equation}
where $p,q=0,\ldots,L-1$ index positions in the lookback window. $\Sigma$ is the covariance matrix of the $L$ past input. $\mathbf{c}=(c_0,\ldots,c_{L-1})^\top$ is a vector that describes how strongly each of the $L$ past input is related to the target target value $y_{t+H}$. Because the past input and the target are jointly Gaussian, the variance remaining after observing the past is given by:
\begin{equation}
\label{appeq:gp}
    \operatorname{Var}(y_{t+H}\mid y_{t-L+1:t})
    =
    \gamma(0)-\mathbf{c}^\top\Sigma^{-1}\mathbf{c}.
\end{equation}
Following~\citet{ranjan2011computationally} and~\citet{terenin2024numerically}, we add $10^{-10}\gamma(0)$ to diagonal of $\Sigma$ for stability.

For each lookback window of $L$, the forward algorithm~\citep{rabiner1989tutorial} estimates the posterior probabilities of the two hidden states at the end of the window. The posterior is propagated $H$ steps through the transition matrix, giving future-state probabilities $\pi_{t+H,i}$ in state $i$ at time $t+H$. Using the Gaussian-mixture variance formula of~\citet{wen2019gaussian}, the predictive variance is given by
\begin{equation}
\label{appeq:hmm}
    \operatorname{Var}(y_{t+H}\mid y_{t-L+1:t})
    =
    \sum^2_{i=1} \pi_{t+H,i}(\mu_i^2+\sigma_i^2)
    -
    \left(\sum^2_{i=1} \pi_{t+H,i}\mu_i\right)^2.
\end{equation}
We average the predictive variance over $400$ lookback windows and use the result to calculate the reference predictability score $P(L,H)$.

For the noisy Lorenz system,  we generate $500$ possible hidden Lorenz states for each lookback window from its target values using a bootstrap particle filter~\citep{gordon1993novel,kitagawa1996monte}. We then evolve these states under the known Lorenz dynamics and measure how much their future target values differ. Since the target is the first state variable plus Gaussian noise, the predictive variance is estimated as
\begin{equation}
\label{appeq:nl}
    \operatorname{Var}(y_{t+H}\mid y_{t-L+1:t})
    \approx
    \operatorname{Var}(z_{t+H,1}\mid y_{t-L+1:t})
    +\sigma_v^2.
\end{equation}
Here, $\operatorname{Var}$ is the variance estimated from the propagated particles. We average this variance over $400$ lookback windows for obtaining $P(L,H)$.

\subsection{Construction of the \texttt{ILI} datasets}
\label{appsubsec:ili_data}

Following~\citet{shu2025deformtime}, we construct daily influenza-like illness (ILI) forecasting datasets for England and US Health and Human Services Regions 2, 9, and additionally included region 10. The forecasting target is the ILI rate for each region. ILI rates are obtained from the Royal College of General Practitioners (RCGP) for England and from the Centers for Disease Control and Prevention (CDC) for the US regions. Since the original ILI rates are reported weekly, we linearly interpolate them to a daily resolution. We use daily frequencies of health-related Google search trends as exogenous inputs. For the US regions, search frequencies are collected at the state level and combined to obtain a regional time series. Additionally, we consider the lag in reported ILI rates, specifically $7$ days for England and $14$ days for the US regions, following the setup in DeformTime~\citep{shu2025deformtime}.

\subsection{Selection of prediction regions from the WeatherBench dataset}
\label{appsubsec:weather_data}

\begin{figure}[t]
    \centering
    \includegraphics[width=0.85\linewidth]{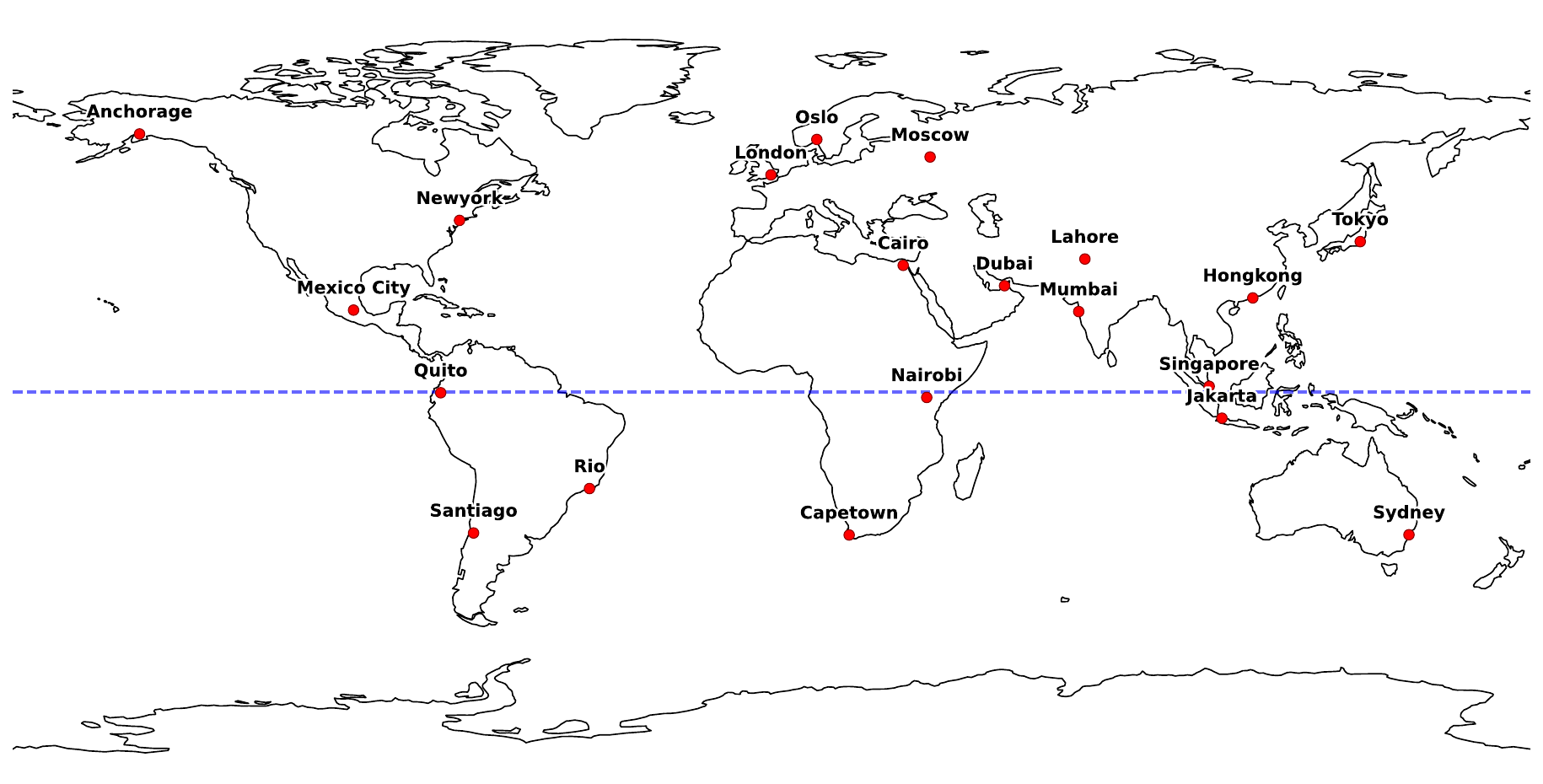}
    \caption{Distribution of the 20 locations used in the weather forecasting tasks.}
    \label{fig:20global_cities}
\end{figure}

We form the weather forecasting (\texttt{WEA}) dataset using climate information extracted from the WeatherBench repository~\citep{rasp2020weatherbench}, covering the period from 1979 to 2018, following the data collection process of Sonnet~\citep{shu2026sonnet}. Data is sampled from a uniform spatial grid over the globe with a resolution of $5.625^\circ$ in both latitude and longitude, where each grid point represents a geographical location. To capture a broader range of climatic conditions, we select $20$ spatially distant locations distributed across different continents, hemispheres, climate zones, and elevations, as labelled in~\figureautorefname~\ref{fig:20global_cities}. This allows the dataset to cover diverse weather patterns. For each location, climate data is sampled from the grid point closest to its geographical coordinates. The forecasting target is the 850 hPa temperature in Kelvin (T850), a widely used indicator of atmospheric conditions. Following previous work~\citep{shu2026sonnet}, we include $4$ additional climate variables as exogenous inputs, namely the 500 hPa geopotential (Z500), 2-metre air temperature (t2m), 10-metre zonal wind (u10), and 10-metre meridional wind (v10). The data is resampled to a temporal resolution of 6 hours. To investigate the contribution of spatial context, we construct two variants for each location. The first contains 45 covariates. It includes all $5$ above-mentioned variables from the $8$ neighbouring grid points surrounding the target location. This forms a $3 \times 3$ spatial window centred on the selected grid point, resulting in 45 input variables. The second contains $5$ local climate variables at the target grid point, without accounting for the neighbouring grid points. 

\section{Supplementary Experiment Setup}
\label{appsec:supplementary_exp}

\subsection{Forecasting setup and data splits}
Models are trained using Adam with a learning rate of $2\times10^{-4}$. We set the maximum training epochs to $100$ with early stopping based on validation loss, with a patience of $5$ epochs. Following Sonnet~\citep{shu2026sonnet} for \texttt{WEA}, years 2016, 2017, and 2018 are used for testing, with the preceding year used for validation and all earlier data (since 1980) used for training. Following DeformTime~\citep{shu2025deformtime} for \texttt{ILI}, we use $4$ test seasons from 2015/16 to 2018/19. The preceding $9$ seasons are used for training, and three $60$-day periods from the most recent training seasons are selected as a held-out set for validation. All inputs are standardised using training-set statistics. For each task, the predictability score is computed on the training set, while the forecasting performance is evaluated on the test set. All experiments are conducted using a server running Ubuntu 24.04.4, with 2 AMD EPYC 9354 CPUs, 768GB of DDR5 RAM, and 3 NVIDIA L40S GPUs with 48GB VRAM each.

\subsection{Details of the ablation components with variance}
\label{appsubsec:abltaion_components}

We describe the variance-based component evaluated in the ablation study in Section~\ref{subsec:ablation}, which replaces the direction-based similarity measure used by~\oursShort{}. Specifically, given the neighbourhood $\mathcal{N}_t$ defined in Section~\ref{subsec:data_pred}, we compute:
\begin{equation}
    v_t
    =
    \frac{1}{H}
    \sum_{h=1}^{H}
    \frac{1}{k-1}
    \sum_{j\in\mathcal{N}_t}
    \left(y_{j,h}-\bar{y}_{t,h}\right)^2,
    \qquad
    \bar{y}_{t,h}
    =
    \frac{1}{k}
    \sum_{j\in\mathcal{N}_t} y_{j,h}.
\end{equation}
Here, $y_{j,h}$ denotes the $h$-th element of the standardised target vector $\mathbf{y}_j$, and $\bar{y}_{t,h}$ is its mean across the neighbours of window $t$. Thus, $v_t$ measures the variance of the neighbouring targets averaged across the $H$ forecast steps. We define the window-level predictability as
\begin{equation}
u_t
=
1-\min(v_t,1) \, \, ,
\end{equation}
and the overall variance-based score as
\begin{equation}
\mathcal{P}_{\mathrm{var}}
=
\frac{1}{N}\sum_{t=1}^{N}u_t.
\end{equation}
A smaller $v_t$ indicates that neighbouring input windows have more clustered future targets and therefore yielding a higher predictability score.

\subsection{Summary of baseline predictability methods}
\label{appsubsec:baseline_pred}

We compare against $6$ baseline methods for dataset predictability.

\textbf{Persistence (Na\"ive).} The persistence baseline predicts target values with the last available observation. We thereby define persistence predictability as the reduction in absolute error relative to predicting the mean target value:
$$
P
=1-\frac{\sum_{t=1}^{N}\sum_{h=1}^{H}| y_{t+h}-y_{t}|}
{\sum_{t=1}^{N}\sum_{h=1}^{H}| y_{t+h}-\bar y|},
\qquad
\bar y=\frac{1}{NH}\sum_{t=1}^{N}\sum_{h=1}^{H} y_{t+h}.
$$

\textbf{Forecastable Component Analysis (ForeCA).} Proposed by~\citet{goerg2013forecastable}, ForeCA defines the predictability of a univariate time series as the complement of the normalised Shannon entropy of its power spectral density. We apply this measure to the complete target series for each training split and report the average across splits.\footnote{Our implementation uses the \href{https://cran.r-project.org/web/packages/ForeCA/index.html}{ForeCA package}.}

\textbf{Spectral predictability.} Proposed by~\citet{wang2025spectral}, this method quantifies predictability as one minus the normalised spectral entropy of the power spectrum of a univariate time series. For each training split, we compute the score over the complete series of each covariate and the target separately and average the resulting scores across variables. We then report the average across training splits.

\textbf{Pattern Complexity (PC).} Proposed by~\citet{wang2025accuracy}, PC measures the variation in frequency-domain patterns across forecasting windows, with larger values indicating greater complexity. For each input variable, including the target history, we concatenate its $L$-step lookback window with the corresponding $H$-step future target, compute PC across the resulting windows, and average the scores across variables. We then report the average across training splits.

\textbf{Mutual Information (MI).} Motivated by the predictability measurement proposed in~\citet{delsole2004predictability}, we adapt mutual information to multivariable forecasting tasks by constructing paired samples from rolling windows, with an $L$-step look-back window as input and an $H$-step target horizon as output. We estimate MI for each input feature using \citet{kraskov2004estimating} and report the average score.

\textbf{Epiplexity.} Following \citet{finzi2026entropy}, epiplexity measures the learnable information in a dataset, estimated as the area between the training loss curve and the final loss. We approximate epiplexity using ridge regression as a baseline measure of dataset predictability, with higher values indicating more learnable structure. Notably, since epiplexity is model-dependent, it may yield different values under different architectures.
\section{Supplementary Results}

\subsection{Results with synthetic data}
\label{appsubsec:synthetic_results}

\begin{figure}
    \centering
    \includegraphics[width=0.95\linewidth]{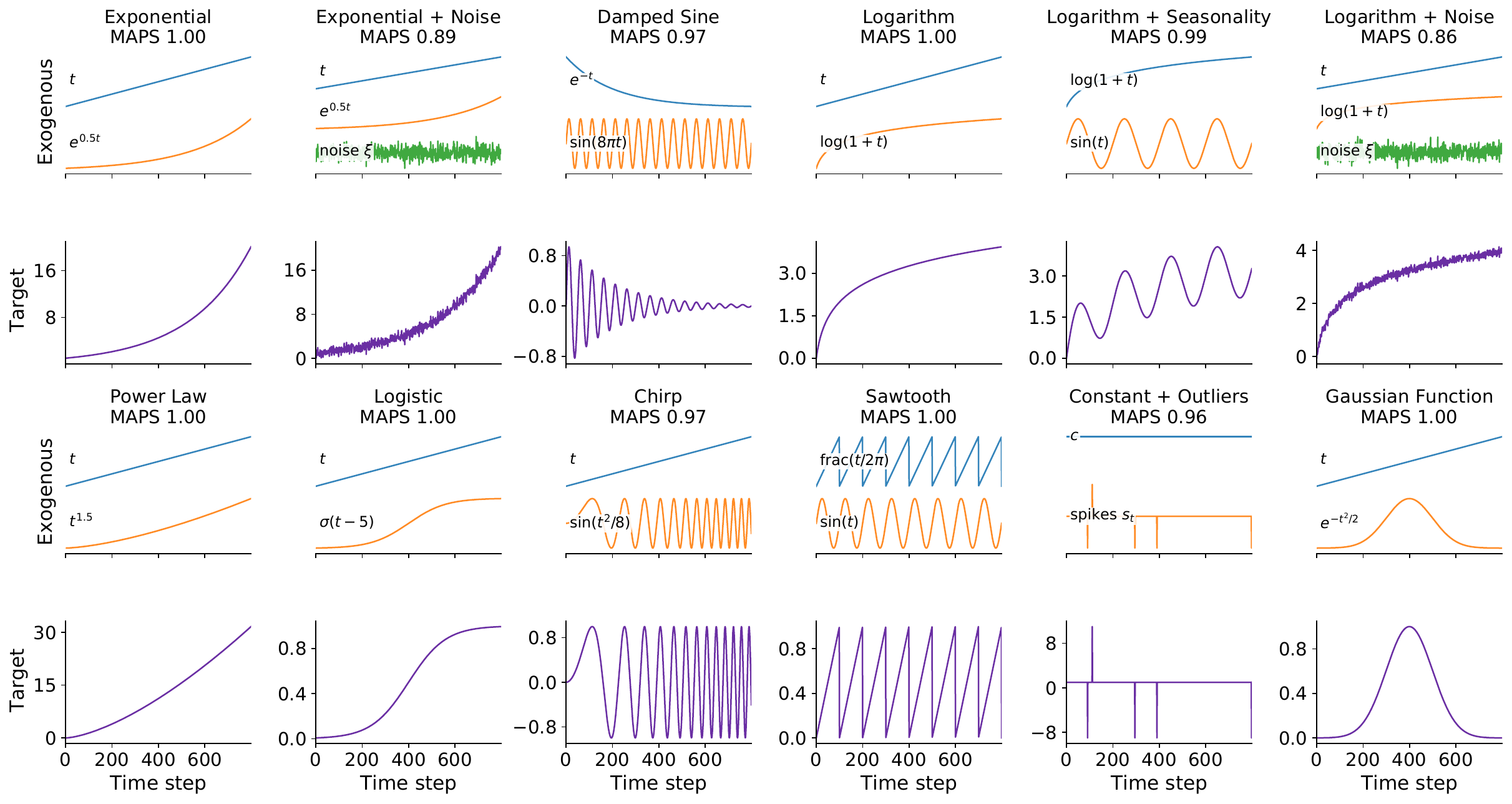}
    \caption{Additional synthetic data and corresponding \oursShort{} scores. Exogenous variables are shown in the first and third rows, with the corresponding target series directly below.}
    \label{appfig:additional_synthetic}
\end{figure}

\begin{table*}[!b]
    \centering
    \caption{Predictability scores of bounded baseline methods on additional synthetic time series. All scores lie in $[0,1]$, with higher values indicating greater estimated predictability.}
    \label{apptab:synthetic_function_baselines}
    \setlength{\tabcolsep}{4pt}
    \setlength{\aboverulesep}{1.2pt}
    \setlength{\belowrulesep}{1.2pt}
    \small
    \resizebox{\textwidth}{!}{%
        \begin{tabular}{lcccccccccccc}
            \toprule
            Method
            & Exponential
            & Exponential + Noise
            & Damped Sine
            & Logarithm
            & Log + Seasonal
            & Log + Noise
            & Power Law
            & Logistic
            & Chirp
            & Sawtooth
            & Constant + Outliers
            & Gaussian Function \\
            \midrule
            \oursShort{}
            & 0.998 & 0.888 & 0.965 & 0.998 & 0.990 & 0.864
            & 0.998 & 0.999 & 0.975 & 0.996 & 0.965 & 0.997 \\
            ForeCA
            & 0.453 & 0.100 & 0.463 & 0.453 & 0.455 & 0.101
            & 0.453 & 0.453 & 0.461 & 0.453 & 0.789 & 0.453 \\
            Spectral
            & 0.919 & 0.520 & 0.918 & 0.919 & 0.919 & 0.521
            & 0.919 & 0.919 & 0.918 & 0.916 & 0.898 & 0.919 \\
            MI
            & 1.000 & 0.580 & 0.268 & 1.000 & 0.756 & 0.562
            & 1.000 & 1.000 & 0.556 & 1.000 & 0.003 & 1.000 \\
            \bottomrule
        \end{tabular}%
    }
\end{table*}

To validate whether~\oursShort{} captures predictability across series with broader patterns, we construct additional synthetic data as shown in Figure~\ref{appfig:additional_synthetic}, with detailed results reported in Table~\ref{apptab:synthetic_function_baselines}. We show rows of exogenous-variable plots, each followed by the corresponding target-variable plots. Each dataset contains one or more exogenous variables, with the target sequence generated as a known function of the inputs. These data include exponential, logarithmic, and periodic patterns such as the chirp, sawtooth, \etc. It can be observed that~\oursShort{} assigns scores close to 1 across these deterministic systems. In comparison, Spectral produces consistently high scores of approximately $0.92$, while ForeCA remains near $0.45$. MI correctly identifies the monotonic deterministic functions but assigns substantially lower scores to the damped sine and chirp signals ($0.27$ and $0.56$, respectively). Additionally, we include a constant series with sparse outlier spikes. This series receives a high~\oursShort{} score of $0.97$. This result indicates that infrequent irregularities have a limited effect on the overall predictability of the series. Since the underlying series remains largely constant, the outliers introduce limited uncertainty relative to the overall signal.

\begin{table}[!t]
    \centering
    \begin{minipage}[t]{0.47\linewidth}
    \centering
    \caption{Subtraction-corrected \oursShort{} predictability score analysis \wrt different neighbourhood size $k$, averaged over $20$ locations.}
    \label{apptab:k_scores}
    \scriptsize
    \setlength{\tabcolsep}{2.0pt}
    \resizebox{\textwidth}{!}{%
    \begin{tabular}{llccccccccc}
    \toprule
        \multirow{2}{*}{$k$} 
        & $L$ & \multicolumn{3}{c}{$28$} & \multicolumn{3}{c}{$56$} & \multicolumn{3}{c}{$84$} \\
        \cmidrule(lr){3-5}\cmidrule(lr){6-8}\cmidrule(lr){9-11}
        & $H$ & $4$ & $12$ & $28$ & $4$ & $12$ & $28$ & $4$ & $12$ & $28$ \\
    \midrule
        1 & & .031 & .045 & .052 & .026 & .036 & .045 & .025 & .036 & .044 \\
        2 & & .283 & .255 & .237 & .264 & .243 & .229 & .259 & .238 & .228 \\
        8 & & .433 & .396 & .367 & .411 & .381 & .359 & .401 & .373 & .356 \\
        12 & & .464 & .424 & .393 & .442 & .409 & .384 & .432 & .401 & .382 \\
        32 & & .513 & .468 & .434 & .492 & .454 & .425 & .484 & .447 & .423 \\
        48 & & .525 & .479 & .443 & .504 & .465 & .435 & .496 & .459 & .433 \\
        64 & & \textbf{.530} & \textbf{.483} & \textbf{.448} & .510 & .471 & .440 & .503 & .465 & .438 \\
        256 & & .529 & .482 & .447 & \textbf{.514} & \textbf{.474} & \textbf{.442} & \textbf{.508} & \textbf{.469} & \textbf{.441} \\
        1024 & & .480 & .440 & .408 & .473 & .437 & .408 & .468 & .432 & .404 \\
        2048 & & .416 & .382 & .354 & .414 & .383 & .357 & .411 & .379 & .355 \\
        4096 & & .185 & .169 & .155 & .184 & .170 & .158 & .184 & .169 & .157 \\
        all & & .000 & .000 & .000 & .000 & .000 & .000 & .000 & .000 & .000 \\
    \bottomrule
    \end{tabular}
    }%
    \end{minipage}
    \hfill
    \begin{minipage}[t]{0.47\linewidth}
    \centering
    \caption{Linear correlation between~\oursShort{} score and Sonnet's forecasting performance, taken across the $20$ regions within each setting.}
    \label{apptab:k_scores_r}
    \scriptsize
    \setlength{\tabcolsep}{2.0pt}
    \resizebox{\textwidth}{!}{%
    \begin{tabular}{llccccccccc}
    \toprule
        \multirow{2}{*}{$k$} 
        & $L$ & \multicolumn{3}{c}{$28$} & \multicolumn{3}{c}{$56$} & \multicolumn{3}{c}{$84$} \\
        % \cmidrule(lr){2-4}\cmidrule(lr){5-7}\cmidrule(lr){8-10}
        \cmidrule(lr){3-5}\cmidrule(lr){6-8}\cmidrule(lr){9-11}
         % & $H=4$ & $H=12$ & $H=28$ & $H=4$ & $H=12$ & $H=28$ & $H=4$ & $H=12$ & $H=28$ \\
         & $H$ & $4$ & $12$ & $28$ & $4$ & $12$ & $28$ & $4$ & $12$ & $28$ \\
    \midrule
        1 & & \textbf{.942} & .943 & .937 & \textbf{.932} & .958 & .950 & \textbf{.943} & .954 & .948 \\
        2 & & .879 & .971 & .958 & .849 & .976 & .966 & .836 & .976 & .967 \\
        8 & & .856 & .976 & .967 & .820 & .977 & .972 & .804 & .976 & .973 \\
        12 & & .852 & .976 & .968 & .815 & .977 & .972 & .798 & .976 & .973 \\
        32 & & .846 & .977 & .970 & .810 & .977 & .973 & .793 & .976 & .975 \\
        48 & & .844 & .977 & .971 & .808 & .977 & .974 & .794 & .976 & .976 \\
        64 & & .843 & \textbf{.978} & .971 & .809 & .978 & .975 & .794 & .976 & .976 \\
        256 & & .845 & .976 & \textbf{.972} & .820 & \textbf{.980} & \textbf{.979} & .807 & \textbf{.977} & \textbf{.980} \\
        1024 & & .840 & .954 & .956 & .827 & .967 & .969 & .824 & .961 & .972 \\
        2048 & & .803 & .903 & .912 & .799 & .921 & .928 & .803 & .913 & .935 \\
        4096 & & .764 & .837 & .853 & .748 & .853 & .865 & .747 & .837 & .872 \\
        all & & .360 & .424 & .433 & .342 & .428 & .441 & .339 & .430 & .415 \\
    \bottomrule
    \end{tabular}
    }%
    \end{minipage}
\end{table}

\subsection{Further analysis on the number of nearest neighbours ($k$) in MAPS}

We use the \texttt{WEA} dataset to assess how the neighbourhood size $k$ affects the predictability score. Results are enumerated in~\tableautorefname~\ref{apptab:k_scores}. For each of the $20$ weather regions and $9$ forecasting tasks, we evaluate the effect for values of $k$ ranging from $1$ to all available neighbours and average the results across regions. To compare results across different values of $k$, we remove the similarity that occurs by chance. Specifically, we randomly shuffle the target vectors $\mathbf{y}_t$ across the input samples while keeping the input vectors $\mathbf{z}_t$ and their neighbourhoods $\mathcal{N}_t$ unchanged. We then recompute the predictability score using the shuffled targets and report $\mathcal{P}_{\mathrm{subtraction}}=\mathcal{P}(\mathbf{X}\rightarrow\mathbf{y})-\mathcal{P}(\mathbf{X}\rightarrow\mathbf{y}_{\mathrm{shuffled}})$. The shuffled score measures the neighbourhood similarity expected when the inputs and their futures are unrelated. Subtracting it removes the chance similarity associated with each value of $k$.

This subtraction is particularly important for small values of $k$. When $k=1$, each input window is evaluated only on the target vector of its nearest input neighbour. Increasing $k$ provides a more stable estimate of local future behaviour by combining similarity scores from multiple nearby inputs. At $k=8$, the original~\oursShort{} score exceeds the permutation baseline by at least $0.35$ across all forecasting tasks, showing that it captures substantially more structure than $\mathcal{P}_{\mathrm{subtraction}}$. Overall, intermediate values of $k$ provide the clearest predictability signal: overly small neighbourhoods offer limited evidence beyond chance, whereas very large neighbourhoods dilute the local relationship between inputs and their futures.

We also examine how well~\oursShort{} reflects forecasting performance across the same range of $k$ in~\tableautorefname~\ref{apptab:k_scores_r}. Specifically, we compute the linear correlation $r$ between~\oursShort{} and Sonnet's forecasting performance $\rho$. The correlation remains high across a broad range of $k$, including our default choice $k=\lfloor\log_2(N)\rfloor=12$. Interestingly, the three $H=4$ tasks reach the highest correlation at $k=1$, likely because the target prediction of each input sample is compared only with that of its closest neighbour, preserving small differences in short-term dynamics. On the other hand, short-horizon targets are also easier to predict, reflected in the forecasting performance. Meanwhile, most longer-horizon settings peak at moderate values of $k$. This indicates that the default setting of $k$ provides a robust choice without requiring $k$ to be tuned separately for each task.

\subsection{Results with additional metrics}
\label{appsubsec:results_mae_smape}

\begin{figure}[!t]
    \centering
    \includegraphics[width=0.95\linewidth]{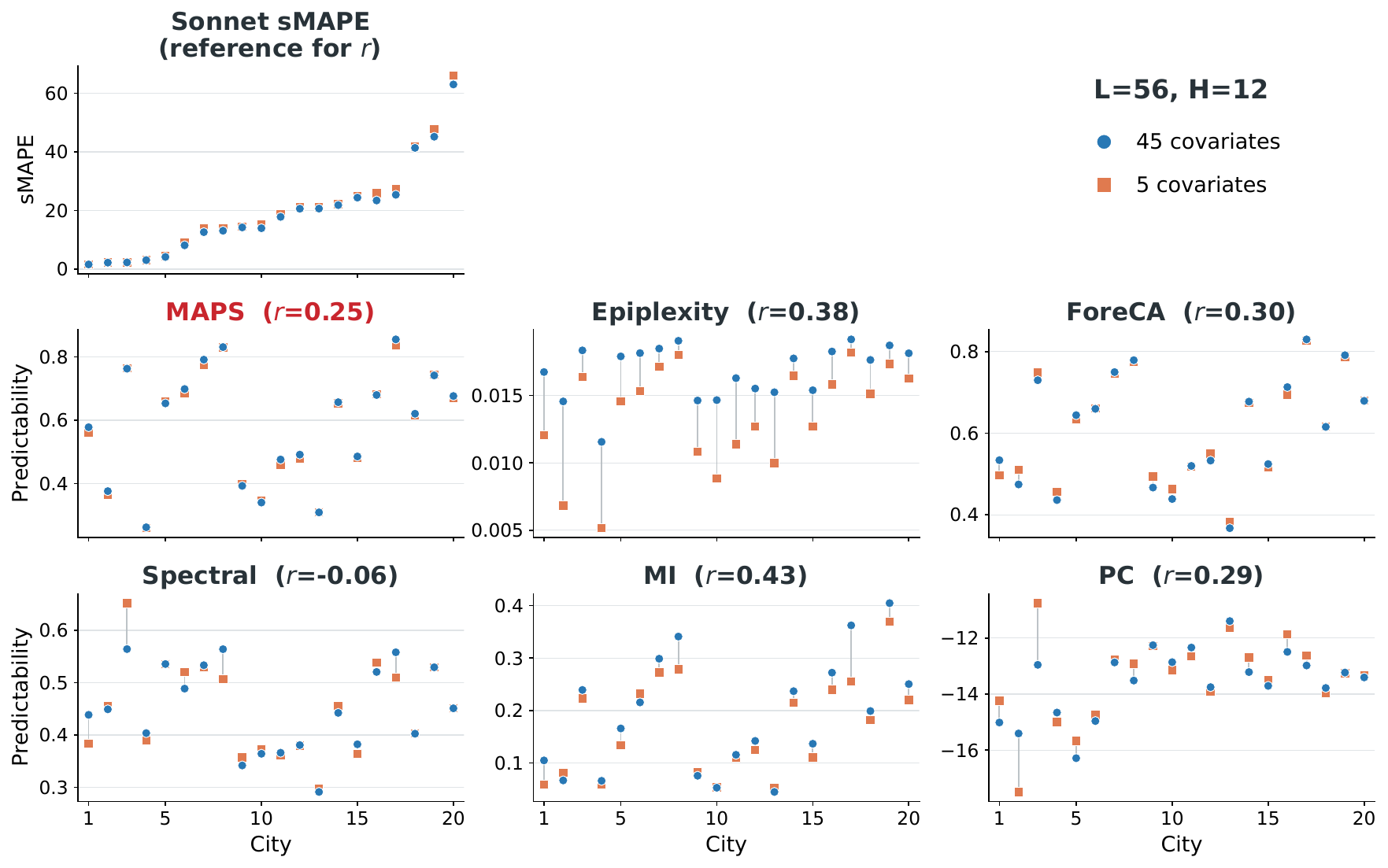}
    \caption{Relationship between different predictability measures and the target SMAPE across all evaluated locations under the Sonnet forecasting model.}
    \label{appfig:smape_sonnet}
\end{figure}

\figureautorefname~\ref{appfig:smape_sonnet} compares the predictability scores with Sonnet's SMAPE. For the reasons discussed in Section~\ref{appsubsec:metrics_choice}, we report these results for completeness rather than use SMAPE as the primary reference. It can be observed that all methods correlate poorly with SMAPE, and stronger alignment with SMAPE does not necessarily indicate a more accurate predictability measure.

\subsection{Results with additional machine learning model}
\label{appsubsec:results_crossformer}

To verify that the analysis of~\oursShort{} is not limited to a single forecasting model, we further provide results with Crossformer~\cite{zhang2023crossformer} in place of Sonnet. \tableautorefname~\ref{apptab:weather_crossformer_correlations} reports the linear correlation $r$ across $20$ locations between each predictability score and Crossformer's performance, measured by $\rho$ and $s$. When comparing the predictability scores with Crossformer's target correlation $\rho$, \oursShort{} yields the strongest correlation $r$ in $8$ of the $18$ configurations. It performs best in five of the six configurations with $H=12$ and three of the six configurations with $H=28$. Epiplexity is the most competitive baseline, leading all six configurations with $H=4$ and three configurations with $H=28$, while ForeCA leads the remaining $H=12$ configuration. Turning to the scaled MAE score $s$, \oursShort{} yields the strongest correlation in $13$ of the $18$ configurations, including five of the six configurations at both $H=12$ and $H=28$. ForeCA leads the remaining five configurations. These results show that the strong alignment between \oursShort{} and forecasting performance is not limited to Sonnet.

\begin{table*}[!b]
\centering
\small
\setlength{\tabcolsep}{3.2pt}
\caption{Linear correlation $r$ between predictability scores and Crossformer's forecasting performance, measured by target correlation $\rho$ or scaled MAE $s$, across the $20$ \texttt{WEA} locations. Predictability scores and model performance are averaged over the 2016--2018 test periods before computing $r$. Results are reported across lookback windows ($L$), forecasting horizons ($H$), and covariate numbers ($C$). The best results are bolded.}
\label{apptab:weather_crossformer_correlations}
\resizebox{\textwidth}{!}{%
\begin{tabular}{cl*{18}{c}}
\toprule
& \multirow{3}{*}{Method}
& \multicolumn{6}{c}{$L=28$}
& \multicolumn{6}{c}{$L=56$}
& \multicolumn{6}{c}{$L=84$} \\
\cmidrule(lr){3-8}
\cmidrule(lr){9-14}
\cmidrule(lr){15-20}
&
& \multicolumn{2}{c}{$H=4$}
& \multicolumn{2}{c}{$H=12$}
& \multicolumn{2}{c}{$H=28$}
& \multicolumn{2}{c}{$H=4$}
& \multicolumn{2}{c}{$H=12$}
& \multicolumn{2}{c}{$H=28$}
& \multicolumn{2}{c}{$H=4$}
& \multicolumn{2}{c}{$H=12$}
& \multicolumn{2}{c}{$H=28$} \\
\cmidrule(lr){3-4}
\cmidrule(lr){5-6}
\cmidrule(lr){7-8}
\cmidrule(lr){9-10}
\cmidrule(lr){11-12}
\cmidrule(lr){13-14}
\cmidrule(lr){15-16}
\cmidrule(lr){17-18}
\cmidrule(lr){19-20}
&
& $C=45$ & $C=5$
& $C=45$ & $C=5$
& $C=45$ & $C=5$
& $C=45$ & $C=5$
& $C=45$ & $C=5$
& $C=45$ & $C=5$
& $C=45$ & $C=5$
& $C=45$ & $C=5$
& $C=45$ & $C=5$ \\
\midrule

\multirow{7}{*}{\rotatebox[origin=c]{90}{$\rho$}}
& \oursShort{}
& .847 & .898 & \textbf{.973} & .966
& \textbf{.970} & .965
& .811 & .858 & \textbf{.977} & \textbf{.972}
& \textbf{.971} & .970
& .800 & .847 & \textbf{.974} & \textbf{.975}
& \textbf{.970} & .969 \\

& Epiplexity
& \textbf{.989} & \textbf{.959} & .957 & .947
& .968 & \textbf{.970}
& \textbf{.988} & \textbf{.953} & .926 & .945
& .943 & \textbf{.971}
& \textbf{.958} & \textbf{.956} & .860 & .951
& .858 & \textbf{.971} \\

& ForeCA
& .805 & .838 & .972 & \textbf{.969}
& .967 & .968
& .803 & .829 & .975 & .970
& .969 & .967
& .810 & .841 & .972 & .973
& .964 & .966 \\

& Spectral
& .467 & .513 & .831 & .764
& .850 & .774
& .463 & .506 & .837 & .775
& .853 & .785
& .477 & .516 & .843 & .774
& .869 & .786 \\

& MI
& .746 & .782 & .901 & .864
& .910 & .880
& .718 & .750 & .904 & .870
& .922 & .884
& .718 & .750 & .903 & .877
& .922 & .883 \\

& PC
& .348 & .461 & .041 & .311
& -.017 & .301
& .311 & .452 & -.065 & .222
& -.070 & .219
& .297 & .453 & -.118 & .206
& -.142 & .193 \\

% & Persistence
% & .891 & .902 & .918 & .913
% & .936 & .942
% & .885 & .902 & .919 & .914
% & .932 & .941
% & .886 & .900 & .914 & .907
% & .927 & .932 \\

\midrule

\multirow{7}{*}{\rotatebox[origin=c]{90}{$s$}}
& \oursShort{}
& \textbf{.924} & \textbf{.944} & \textbf{.961} & .946
& \textbf{.955} & .949
& .913 & \textbf{.934} & \textbf{.965} & \textbf{.952}
& \textbf{.955} & \textbf{.952}
& .905 & .929 & \textbf{.968} & \textbf{.959}
& \textbf{.950} & \textbf{.956} \\

& Epiplexity
& .918 & .904 & .911 & .893
& .938 & .935
& .913 & .902 & .882 & .886
& .915 & .936
& .876 & .903 & .832 & .891
& .836 & .937 \\

& ForeCA
& .916 & .932 & .958 & \textbf{.954}
& .950 & \textbf{.950}
& \textbf{.920} & .931 & .958 & .951
& .950 & .949
& \textbf{.920} & \textbf{.935} & .957 & .955
& .940 & .951 \\

& Spectral
& .680 & .665 & .904 & .811
& .898 & .825
& .681 & .666 & .905 & .818
& .898 & .826
& .692 & .668 & .910 & .821
& .904 & .827 \\

& MI
& .898 & .885 & .911 & .856
& .909 & .871
& .888 & .876 & .920 & .865
& .924 & .879
& .885 & .871 & .920 & .877
& .919 & .883 \\

& PC
& .318 & .413 & .042 & .281
& -.007 & .253
& .215 & .349 & -.094 & .175
& -.079 & .176
& .161 & .315 & -.156 & .140
& -.145 & .149 \\

% & Persistence
% & .918 & .912 & .899 & .896
% & .929 & .933
% & .917 & .918 & .901 & .892
% & .922 & .929
% & .913 & .911 & .893 & .884
% & .910 & .924 \\

\bottomrule
\end{tabular}%
}
\end{table*}

\subsection{Additional analysis on predictability with univariate setup}
\label{appsubsec:predictability_univariate}

\begin{table*}[t]
\centering
\small
\setlength{\tabcolsep}{3.0pt}
\caption{Linear correlation $r$ between target-only predictability scores and Sonnet's forecasting performance. Arrows indicate whether $r$ increases or decreases relative to the corresponding result in~\tableautorefname~\ref{tab:weather_sonnet_correlations} and $=$ indicates no change.}
\label{apptab:correlations_univariate}
\resizebox{\textwidth}{!}{%
\begin{tabular}{cl*{18}{c}}
\toprule
& \multirow{3}{*}{Method}
& \multicolumn{6}{c}{$L=28$}
& \multicolumn{6}{c}{$L=56$}
& \multicolumn{6}{c}{$L=84$} \\
\cmidrule(lr){3-8}
\cmidrule(lr){9-14}
\cmidrule(lr){15-20}
&
& \multicolumn{2}{c}{$H\!=\!4$}
& \multicolumn{2}{c}{$H\!=\!12$}
& \multicolumn{2}{c}{$H\!=\!28$}
& \multicolumn{2}{c}{$H\!=\!4$}
& \multicolumn{2}{c}{$H\!=\!12$}
& \multicolumn{2}{c}{$H\!=\!28$}
& \multicolumn{2}{c}{$H\!=\!4$}
& \multicolumn{2}{c}{$H\!=\!12$}
& \multicolumn{2}{c}{$H\!=\!28$} \\
\cmidrule(lr){3-4}
\cmidrule(lr){5-6}
\cmidrule(lr){7-8}
\cmidrule(lr){9-10}
\cmidrule(lr){11-12}
\cmidrule(lr){13-14}
\cmidrule(lr){15-16}
\cmidrule(lr){17-18}
\cmidrule(lr){19-20}
&
& $C\!=\!45$ & $C\!=\!5$
& $C\!=\!45$ & $C\!=\!5$
& $C\!=\!45$ & $C\!=\!5$
& $C\!=\!45$ & $C\!=\!5$
& $C\!=\!45$ & $C\!=\!5$
& $C\!=\!45$ & $C\!=\!5$
& $C\!=\!45$ & $C\!=\!5$
& $C\!=\!45$ & $C\!=\!5$
& $C\!=\!45$ & $C\!=\!5$ \\
\midrule

\multirow{5}{*}{\rotatebox[origin=c]{90}{$\rho$}}
& \oursShort{}
& .822$\downarrow$ & .854$\downarrow$ & .974$\downarrow$ & .964$\downarrow$ & .960$\downarrow$ & .954$\downarrow$
& .787$\downarrow$ & .827$\downarrow$ & .974$\downarrow$ & .963$\downarrow$ & .965$\downarrow$ & .952$\downarrow$
& .764$\downarrow$ & .812$\downarrow$ & .974$\downarrow$ & .965$\downarrow$ & .967$\downarrow$ & .962$\downarrow$ \\

& ForeCA
& .801$\uparrow$ & .836$\uparrow$ & .972$\uparrow$ & .974$\uparrow$ & .961$\downarrow$ & .972$\uparrow$
& .804$\uparrow$ & .845$\uparrow$ & .976$\uparrow$ & .974$\uparrow$ & .970$\uparrow$ & .963$\uparrow$
& .805$\uparrow$ & .850$\uparrow$ & .974$\uparrow$ & .972$\uparrow$ & .973$\downarrow$ & .973$\uparrow$ \\

& Spectral
& .806$\uparrow$ & .840$\uparrow$ & .968$\uparrow$ & .970$\uparrow$ & .954$\uparrow$ & .967$\uparrow$
& .808$\uparrow$ & .850$\uparrow$ & .973$\uparrow$ & .970$\uparrow$ & .963$\uparrow$ & .958$\uparrow$
& .810$\uparrow$ & .854$\uparrow$ & .970$\uparrow$ & .969$\uparrow$ & .968$\uparrow$ & .969$\uparrow$ \\

& MI
& .713$\downarrow$ & .759$\downarrow$ & .940$\uparrow$ & .940$\uparrow$ & .944$\uparrow$ & .953$\uparrow$
& .703$\downarrow$ & .757$\downarrow$ & .934$\uparrow$ & .936$\uparrow$ & .948$\uparrow$ & .948$\uparrow$
& .701$\downarrow$ & .755$\downarrow$ & .931$\uparrow$ & .935$\uparrow$ & .951$\uparrow$ & .956$\uparrow$ \\

& PC
& .154$\downarrow$ & .129$\downarrow$ & -.097$\downarrow$ & -.073$\downarrow$ & -.166$\downarrow$ & -.115$\downarrow$
& .015$\downarrow$ & -.014$\downarrow$ & -.229$\downarrow$ & -.212$\downarrow$ & -.263$\downarrow$ & -.232$\downarrow$
& -.031$\downarrow$ & -.065$\downarrow$ & -.318$\downarrow$ & -.301$\downarrow$ & -.287$\downarrow$ & -.278$\downarrow$ \\

\midrule

\multirow{5}{*}{\rotatebox[origin=c]{90}{$s$}}
& \oursShort{}
& .906$\downarrow$ & .933$\downarrow$ & .968$\uparrow$ & .964$\uparrow$ & .960$\uparrow$ & .955$\uparrow$
& .894$\downarrow$ & .919$\downarrow$ & .973$\uparrow$ & .962$\uparrow$ & .957$\downarrow$ & .955$\downarrow$
& .878$\downarrow$ & .911$\downarrow$ & .967$\uparrow$ & .958$=$ & .956$\downarrow$ & .954$\downarrow$ \\

& ForeCA
& .915$\uparrow$ & .936$\uparrow$ & .956$\downarrow$ & .960$\downarrow$ & .947$\downarrow$ & .952$\downarrow$
& .922$\uparrow$ & .938$\uparrow$ & .965$\downarrow$ & .959$\downarrow$ & .953$\downarrow$ & .955$\downarrow$
& .921$\uparrow$ & .940$\uparrow$ & .958$\downarrow$ & .956$\downarrow$ & .953$\downarrow$ & .954$\downarrow$ \\

& Spectral
& .919$\uparrow$ & .940$\uparrow$ & .954$\uparrow$ & .957$\uparrow$ & .943$\uparrow$ & .950$\uparrow$
& .926$\uparrow$ & .941$\uparrow$ & .964$\uparrow$ & .956$\uparrow$ & .950$\uparrow$ & .953$\uparrow$
& .924$\uparrow$ & .944$\uparrow$ & .956$\uparrow$ & .953$\uparrow$ & .950$\uparrow$ & .953$\uparrow$ \\

& MI
& .888$\downarrow$ & .934$\uparrow$ & .987$\uparrow$ & .988$\uparrow$ & .979$\uparrow$ & .982$\uparrow$
& .889$\downarrow$ & .929$\uparrow$ & .987$\uparrow$ & .986$\uparrow$ & .983$\uparrow$ & .984$\uparrow$
& .886$=$ & .924$\uparrow$ & .984$\uparrow$ & .981$\uparrow$ & .986$\uparrow$ & .986$\uparrow$ \\

& PC
& .176$\downarrow$ & .131$\downarrow$ & -.060$\downarrow$ & -.046$\downarrow$ & -.156$\downarrow$ & -.098$\downarrow$
& -.010$\downarrow$ & -.060$\downarrow$ & -.196$\downarrow$ & -.203$\downarrow$ & -.239$\downarrow$ & -.217$\downarrow$
& -.085$\downarrow$ & -.136$\downarrow$ & -.299$\downarrow$ & -.291$\downarrow$ & -.274$\downarrow$ & -.252$\downarrow$ \\

\bottomrule
\end{tabular}%
}
% \vspace{0.3em}
\end{table*}

We further evaluate all methods in a univariate setting, where only the target series is used as input. The resulting scores are compared with Sonnet's performance using $C=45$ and $C=5$ inputs. This analysis therefore evaluates how much information about forecasting difficulty is lost when the predictability measures exclude the external covariates. The results are presented in~\tableautorefname~\ref{apptab:correlations_univariate}. Compared with the multivariable results in~\tableautorefname~\ref{tab:weather_sonnet_correlations}, the correlation between \oursShort{} and $\rho$ decreases in all $18$ configurations. This suggests that the exogenous covariates provide useful information about forecasting difficulty. Spectral improves in every configuration, MI generally improves at longer horizons, PC performs worse, and ForeCA changes only slightly.

\begin{figure}[!b]
    \centering
    \includegraphics[width=0.75\linewidth]{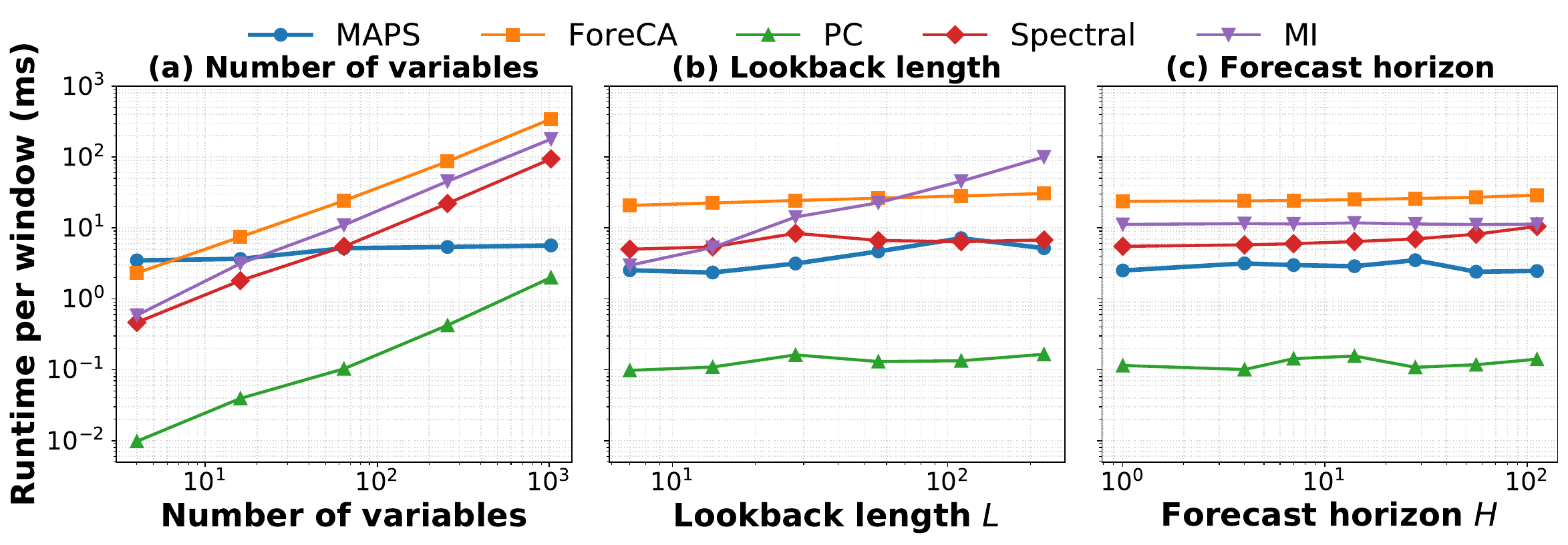}
    \caption{Runtime analysis of predictability estimators. We measure the median per-window runtime (ms) versus the number of variables, lookback window length, and forecast horizon.}
    \label{fig:runtime_comparison}
\end{figure}

\subsection{Runtime Comparison}
\label{appsubsec:runtime}
Figure~\ref{fig:runtime_comparison} compares the median per-window runtime. We change one factor at a time over $C\in\{4,16,64,256,1024\}$, $L\in\{7,14,28,56,112,224\}$, and $H\in\{1,4,7,14,28,56,112\}$, holding the others at $L=28$, $H=7$, and $C=64$. Across the evaluated settings, the runtime of \oursShort{} does not change significantly because its dominant nearest-neighbour search operates in the fixed $d$-dimensional PCA space. By contrast, all baselines become more computationally expensive as the number of variables increases. Additionally, MI also requires more runtime for longer lookback windows.

\section{Limitations}

The main limitation is that \oursShort{} estimates predictability relative to the observed data distribution and should not be interpreted as a theoretical upper bound on forecasting performance. As demonstrated under distribution shift, a score computed on training data may not characterise an unseen deployment distribution. Furthermore, \oursShort{} evaluates directional consistency and does not capture differences in forecast magnitude. Finally, although we evaluate synthetic systems and real-world weather and ILI tasks, further experiments across a broader range of domains are required to establish its general applicability.

\end{document}